\documentclass[lettersize,journal]{IEEEtran}
\usepackage{amsmath,amsfonts}
\usepackage{color}
\usepackage[table,xcdraw]{xcolor}
\usepackage{algorithmic}
\usepackage{algorithm}
\usepackage{array}
\usepackage[caption=false,font=normalsize,labelfont=sf,textfont=sf]{subfig}
\usepackage{textcomp}
\usepackage{stfloats}
\usepackage{url}
\usepackage{verbatim}
\usepackage{graphicx}
\usepackage{cite}
\usepackage{threeparttable}
\usepackage{booktabs}
\usepackage{multirow}
\usepackage{cleveref}
\usepackage{bm}

\usepackage{listings}
\floatstyle{ruled}
\newfloat{listing}{tb}{lst}{}
\floatname{listing}{Listing}
\definecolor{darkgreen}{rgb}{0.0, 0.5, 0.0}
\definecolor{darkred}{rgb}{0.9, 0.0, 0.0}

\begin{document}

\title{COXNet: Cross-Layer Fusion  with Adaptive Alignment and Scale Integration for RGBT Tiny Object Detection}

\author{Peiran Peng, Tingfa Xu, Liqiang Song, Mengqi Zhu, Yuqiang Fang, Jianan Li
\thanks{Corresponding authors: Jianan Li, Tingfa Xu.}

\IEEEcompsocitemizethanks{
\IEEEcompsocthanksitem 
Peiran Peng, Tingfa Xu, Mengqi Zhu and Jianan Li are with Beijing Institute of Technology, Beijing 100081, China (e-mail: \{3220215082, ciom\_xtf1, 3220235095, 3220185049, lijianan\}@bit.edu.cn).
\IEEEcompsocthanksitem 
Liqiang Song is with the National Astronomical Observatories, Chinese Academy of Sciences,Beijing,China (e-mail: lqsong@bao.ac.cn).
\IEEEcompsocthanksitem 
Yuqiang Fang is with the National Key Laboratory of Space Target Awareness, Space Engineering University, Beijing 101416, China.
\IEEEcompsocthanksitem
Jianan Li and Tingfa Xu are also with the Key Laboratory of Photoelectronic Imaging Technology and System, Ministry of Education of China, Beijing 100081, China.
\IEEEcompsocthanksitem
Tingfa Xu is also with Chongqing Innovation Center, Beijing Institute of Technology, Chongqing 401135, China.
\IEEEcompsocthanksitem
Mengqi Zhu is also with China North Vehicle Research Institute, China.
\IEEEcompsocthanksitem
The research is funding by Natural Science Foundation of Chongqing, China (Grant No. cstc2021jcyj-msxmX1130).
}


}

\markboth{Journal of \LaTeX\ Class Files,~Vol.~14, No.~8, August~2021}%
{Shell \MakeLowercase{\textit{et al.}}: A Sample Article Using IEEEtran.cls for IEEE Journals}


\maketitle

\begin{abstract}
Detecting tiny objects in multimodal Red-Green-Blue-Thermal (RGBT) imagery is a critical challenge in computer vision, particularly in surveillance, search and rescue, and autonomous navigation. Drone-based scenarios exacerbate these challenges due to spatial misalignment, low-light conditions, occlusion, and cluttered backgrounds. Current methods struggle to leverage the complementary information between visible and thermal modalities effectively. We propose COXNet, a novel framework for RGBT tiny object detection, addressing these issues through three core innovations: i) the Cross-Layer Fusion Module, fusing high-level visible and low-level thermal features for enhanced semantic and spatial accuracy; ii) the Dynamic Alignment and Scale Refinement module, correcting cross-modal spatial misalignments and preserving multi-scale features; and iii) an optimized label assignment strategy using the GeoShape Similarity Measure for better localization. COXNet achieves a 3.32\% mAP$_{50}$ improvement on the RGBTDronePerson dataset over state-of-the-art methods, demonstrating its effectiveness for robust detection in complex environments. The code is publicly available at \url{https://github.com/Troy-peng-0327/COXNet-release}.
\end{abstract}


\section{Introduction}
Tiny object detection is a crucial computer vision task with applications in surveillance, search and rescue, and autonomous navigation~\cite{xiong2024adaptive}. Detecting tiny objects is inherently difficult due to their small size, which makes them susceptible to background noise. Integrating visible and thermal data in Red-Green-Blue-Thermal (RGBT) imagery can enhance detection, particularly in low visibility or adverse weather~\cite{zhao2022tftn}. However, fusing these modalities is challenging due to differences in spatial resolution, signal characteristics, and modality-specific noise. Fluctuating environmental conditions and background variability, especially in drone-based scenarios, exacerbate these challenges~\cite{10325629, zhang2024high, liao2022cross, sun2022drone, guo2023save, du2023megf}.

\begin{figure}[!t]
    \centering 
    \includegraphics[width=1.0\linewidth]{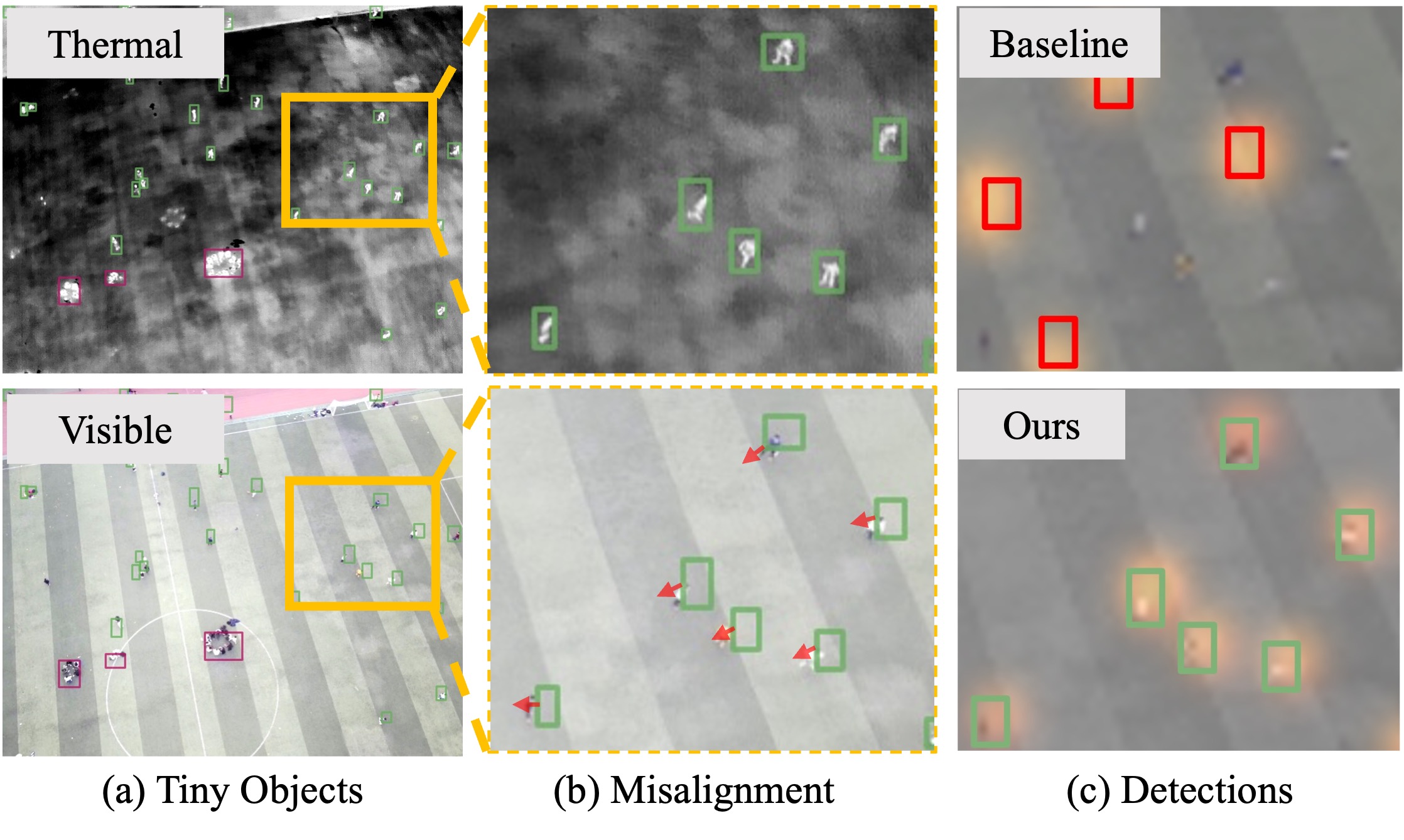}
    \caption{(a)~Challenges in tiny-object detection. \textcolor{orange}{Orange} dashed rectangles highlight pixel-level misalignments between visible and thermal views; \textcolor{darkgreen}{green} and \textcolor{darkred}{red} solid boxes denote the ground-truth locations of persons and crowds, respectively. 
    (b)~Zoom-ins of the \textcolor{orange}{orange} regions in~(a). \textcolor{darkred}{Red} arrows visualize how a few-pixel shift can severely reduce the IoU of tiny boxes. 
    (c)~Qualitative comparison of the {feature maps} inside the zoomed region. \textcolor{orange}{Orange} regions denote strong feature response. \textcolor{darkred}{Red} boxes indicate {false detections or misses produced by the baseline}, whereas \textcolor{darkgreen}{green} boxes mark the correct detections obtained with the proposed COXNet.}
    \label{fig:motivation}
\end{figure}

Drones equipped with RGBT sensors combine visible and thermal information, yet detecting tiny objects remains challenging. Tiny objects occupy a small part of the image, often blending with background clutter. Distinct characteristics of visible and thermal data, such as differences in spatial resolution and signal intensity, complicate fusion and degrade detection performance~\cite{cheng2023cross}. Spatial misalignment between modalities, especially in high-altitude drone imagery, introduces significant deviations~\cite{yuan2022translation}. Misalignments, compounded by occlusion, low-light, and complex backgrounds, further diminish traditional detection performance~\cite{wang2023thermal}. Traditional IoU-based label assignments are too rigid for tiny objects, leading to misclassification from small perturbations~\cite{zhang2023drone}. 

To address these issues in Fig.~\ref{fig:motivation}, we propose COXNet, a novel architecture specifically designed for RGBT tiny object detection in complex environments. COXNet introduces three key innovations to overcome the limitations of existing fusion and alignment methods:

\begin{figure}[ht]
    \centering
    \includegraphics[width=1.0\linewidth]{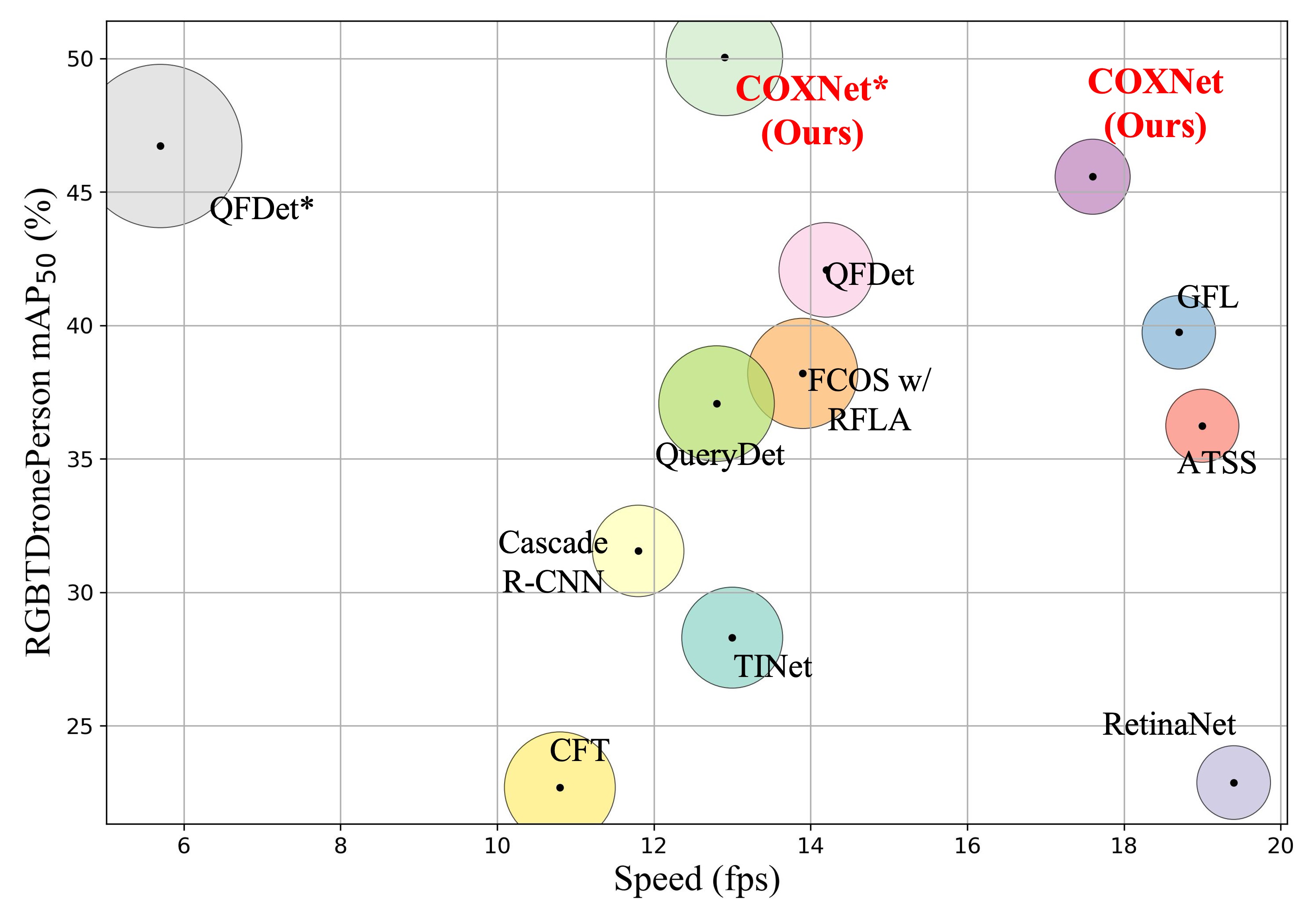}
    \caption{Performance–efficiency trade-off on the RGBTDronePerson dataset. Circle size encodes FLOPs; larger is more expensive. Methods marked with $*$ use a detection head covering P2–P6, while unmarked methods use the standard P3–P7 head.}
    \label{fig:tradeoff_rgbtdroneperson}
\end{figure}

First, we introduce the Cross-Layer Fusion Module (CLFM) to fuse high-level visible and low-level thermal features. Unlike conventional methods that fuse features from similar stages, our approach uses high-level visible features to enhance spatial precision from low-level thermal features. Using wavelet-based alignment, CLFM separates and combines high- an d low-frequency components across modalities. Wavelet transform allows us to decompose features into different frequency bands, facilitating precise alignment and fusion of complementary information from each modality. Compared to traditional alignment techniques, wavelet-based alignment provides a more flexible and robust mechanism for handling cross-modal differences, particularly in terms of spatial resolution and noise characteristics. This approach effectively preserves semantic richness and spatial detail while reducing computational complexity.

Second, we propose the Dynamic Alignment and Scale Refinement (DASR) module to correct spatial misalignment and scale inconsistencies. DASR includes an Adaptive Alignment Module (AAM) for pixel-level alignment, which dynamically adjusts the positions of visible and thermal features to ensure precise spatial correspondence, thereby mitigating cross-modal misalignment. The Dynamic Scale Refinement (DSR) mechanism adjusts feature scales to handle objects of varying sizes, ensuring that both fine-grained details and broader contextual information are captured effectively. This dual mechanism ensures precise alignment and robust multi-scale feature adjustment, capturing both fine-grained and contextual information across scales. Thus, DASR enhances feature representation and enables effective fusion, particularly for drone-based detection under occlusion and variable lighting.

Third, we develop an optimized label assignment using the GeoShape Similarity Measure to improve tiny object localization. Traditional IoU-based label assignment is often inadequate for tiny objects, as it is overly sensitive to small spatial shifts. The GeoShape Similarity Measure captures both spatial and shape characteristics of bounding boxes, ensuring a more robust and adaptive assignment process that improves localization accuracy under challenging conditions.

COXNet is based on the Generalized Focal Loss (GFL)~\cite{li2020generalized} framework, enhancing classification and regression with a Kullback-Leibler (KL) divergence-based loss to ensure cross-modal consistency. This loss explicitly aligns the fused and thermal features, ensuring effective feature representation where it matters most—around tiny objects. The combination of CLFM, DASR, and the GeoShape-based assignment strategy makes COXNet highly effective for tiny object detection in challenging, real-world scenarios.

Experiments on the RGBTDronePerson~\cite{zhang2023drone}, VTUAV-det~\cite{zhang2023drone}, and NII-CU~\cite{speth2022deep} datasets demonstrate that COXNet consistently outperforms state-of-the-art methods across diverse conditions, including low visibility, cluttered backgrounds, and high-altitude drone imaging. As shown in Fig.~\ref{fig:tradeoff_rgbtdroneperson}, the performance-efficiency tradeoff of COXNet on the RGBTDronePerson dataset highlights its superiority in both detection accuracy (mAP$_{50}$), FLOPs and inference speed. COXNet achieves the highest accuracy while maintaining competitive efficiency, striking an optimal balance compared to existing methods. These results underline COXNet’s applicability to real-world scenarios, particularly in resource-constrained environments requiring real-time performance.

Our primary contributions are summarized as follows:
\begin{itemize}
\item We propose the Cross-Layer Fusion Module (CLFM) to fuse high-level visible and low-level thermal features using wavelet-based alignment, enhancing semantic richness and spatial precision.
\item We design the Dynamic Alignment and Scale Refinement (DASR) module, which integrates pixel-level alignment (AAM) and multi-scale feature adjustment (DSR) to address cross-modal misalignment and scale inconsistencies.
\item We develop a GeoShape-based label assignment strategy that improves bounding box localization by capturing spatial and shape similarity, boosting detection performance for tiny objects.
\end{itemize}

\section{Related Work}
\subsection{Multimodal Fusion for Tiny Object Detection}
Multimodal object detection, especially in Red-Green-Blue-Thermal (RGBT) settings, has been extensively studied for various applications, including pedestrian detection in autonomous driving~\cite{li2023multiscale, wang2023dacfn}. Early and middle fusion approaches, like MLPD~\cite{kim2021mlpd}, enhance feature interaction between RGB and thermal data during intermediate network layers, whereas late fusion techniques, such as those in Zhang et al.~\cite{zhang2021weakly}, align features after separate processing to address spatial misalignment challenges. The CFT~\cite{qingyun2021cross} utilizes transformer-based architecture to capture long-range dependencies, improving multispectral feature fusion.

Transformer-based models, such as Damsdet~\cite{guo2025damsdet} and C$^2$Former~\cite{yuan2024c}, leverage self-attention mechanisms to capture global context, thereby boosting the robustness of cross-modal fusion. The Causal Mode Multiplexer~\cite{kim2024causal} uses causal inference to mitigate biases in multispectral pedestrian detection. These models have demonstrated effectiveness for larger objects, particularly in challenging conditions like occlusion or low visibility.

However, existing methods face challenges in tiny object detection, particularly in drone-based scenarios with occlusion and complex backgrounds. DaCFN~\cite{wang2023dacfn} uses a divide-and-conquer approach to separate RGB and thermal features into shared and unique components, but struggles with tiny targets. Similarly, MANBA~\cite{lai2024mambavt} improves fusion efficiency but lacks the ability to effectively capture contextual details needed for tiny object discrimination in cluttered settings.

Our COXNet addresses the challenges of existing methods by introducing modules that correct spatial misalignments and improve cross-modal fusion. Unlike previous approaches, COXNet effectively combines high-level RGB features with low-level thermal features, achieving a better balance between semantic richness, spatial precision, and computational efficiency, making it more suitable for tiny object detection in complex drone-based scenarios.

\subsection{Cross-Layer Fusion Techniques}
Cross-layer fusion techniques are widely used to improve object detection and segmentation, especially in challenging scenarios involving small targets and multimodal data. Recent advancements, such as the TPH-YOLOv5++~\cite{zhao2023tph} framework, introduced cross-layer asymmetric transformers to facilitate efficient feature extraction for drone-based object detection, demonstrating significant improvements for tiny-scale targets while maintaining computational efficiency. QueryDet~\cite{yang2022querydet} employs a cascaded sparse query mechanism to focus computational resources on high-resolution features in regions likely containing small objects, enhancing detection speed while maintaining accuracy. Another method, CFANet~\cite{zhang2023cfanet}, utilizes a context-fusing attentional network that integrates multi-level features using pyramid and scale attention modules to enhance segmentation quality in complex environments involving small targets. Similarly, FAMBA~\cite{shen2024famba} combines multi-scale feature aggregation and cross-layer attention to improve object detection robustness, particularly in cluttered environments.

Our Cross-Layer Fusion Module (CLFM) fuses high-level RGB semantics with low-level thermal features, focusing on preserving spatial accuracy using wavelet transforms for enhanced alignment. This ensures both spatial precision and semantic richness, making CLFM effective for detecting small, occluded objects in UAV scenarios.

\begin{figure*}[ht]
    \centering
    \includegraphics[width=1.0\linewidth]{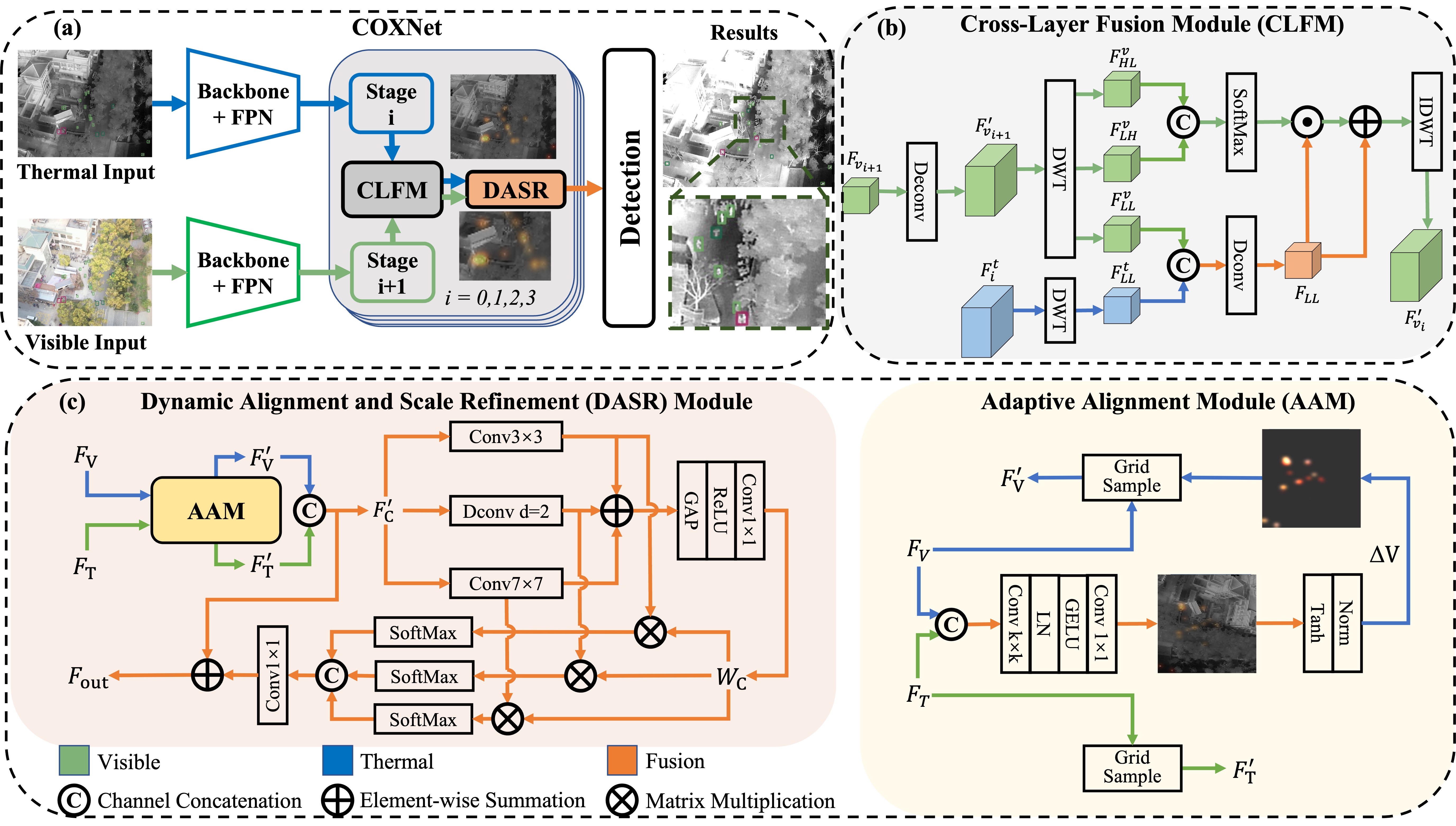}
    \caption{Overall architecture of COXNet. (a) COXNet integrates thermal and visible inputs via independent backbones, employing the \textbf{Cross-Layer Fusion Module (CLFM)} and \textbf{Dynamic Alignment and Scale Refinement (DASR) Module} across multiple stages. (b) CLFM fuses high-level visible and low-level thermal features using wavelet-based decomposition for precise alignment and cross-modal integration. (c) DASR addresses spatial misalignment and scale inconsistencies through the \textbf{Adaptive Alignment Module (AAM)} and multi-kernel convolution, enhancing tiny object detection in challenging conditions.}
    \label{fig:architecture}
\end{figure*}

\subsection{Label Assignment Strategies in RGBT Object Detection}
Label assignment strategies have been pivotal in enhancing multimodal detection by tackling challenges related to modality balance and feature alignment. MBNet~\cite{zhou2020improving} addresses modality imbalance using Differential Modality Aware Fusion and an illumination-aware feature alignment module, effectively balancing cross-modality integration for improved detection under different lighting conditions. Similarly, MLPD~\cite{kim2021mlpd} employs middle fusion to boost the interaction between RGB and thermal features, enhancing coherence across modalities, especially for multispectral pedestrian detection. Moreover, SIWD~\cite{zhang2023drone} introduces a scale-invariant weighted distance metric that provides a more refined mechanism for matching ground truth to predicted bounding boxes, improving the label assignment accuracy for tiny objects. Our proposed GeoShape-based label assignment strategy builds on these methods by explicitly considering both spatial and geometric variations, ensuring more precise localization and alignment between modalities, which is essential for the detection of tiny objects in RGBT imagery.

\section{Method}
COXNet is designed to tackle the challenges of RGBT tiny object detection, including small object size, cross-modal alignment, and cluttered environments. As shown in Fig.~\ref{fig:architecture} (a), the architecture integrates three core components: the Dynamic Alignment and Scale Refinement (DASR) module, the Cross-Layer Fusion Module (CLFM), and an optimized label assignment strategy using the GeoShape Similarity Measure. CLFM combines high-level visible features with low-level thermal features, DASR ensures precise alignment and scale consistency, and the GeoShape-based label assignment strategy improves bounding box accuracy, making COXNet effective for challenging conditions.

\subsection{Cross-Layer Fusion Module}
The Cross-Layer Fusion Module (CLFM) integrates high-level visible features with low-level thermal features to enhance RGBT tiny object detection. As shown in Fig.~\ref{fig:architecture} (b), CLFM consists of two main steps: spatial alignment with wavelet-based decomposition and feature reconstruction through frequency fusion.

\noindent\textbf{Wavelet-Based Spatial Alignment.}
First, the high-level visible feature map, $\bm{F}_{v_{i+1}}$, is spatially aligned with the low-level thermal feature map, $\bm{F}_{t_i}$, by applying deconvolution to adjust its spatial resolution, resulting in $\bm{F}'_{v_{i+1}}$. Unlike direct concatenation in the spatial domain, operating in the frequency domain allows us to decouple \emph{semantic context} (low–frequency LL band) from \emph{fine spatial details} (high–frequency LH/HL/HH bands).  This band–wise separation makes it possible to (i) align and fuse global context while (ii) selectively re-weight high-frequency edges that are critical for tiny objects yet susceptible to modality-specific noise.

The aligned visible feature map, $\bm{F}'_{v_{i+1}}$, and the thermal feature map, $\bm{F}_{t_i}$, are then decomposed using the Discrete Wavelet Transform (DWT):

\begin{equation} 
\bm{F}'_{v_{i+1}} \xrightarrow{\text{DWT}} { \bm{F}^{v}_{{LL}}, \bm{F}^v_{{LH}}, \bm{F}^v_{{HL}}, \bm{F}^v_{{HH}}}, 
\end{equation}

\begin{equation} 
\bm{F}_{t_{i}} \xrightarrow{\text{DWT}} { \bm{F}^{t}_{{LL}}, \bm{F}^t_{{LH}}, \bm{F}^t_{{HL}}, \bm{F}^t_{{HH}}}, 
\end{equation}

The low-frequency components from both modalities, $\bm{F}^v_{LL}$ and $\bm{F}^t_{LL}$, are concatenated and then processed using a dilated convolution to generate the fused low-frequency component, denoted as $\bm{F}_{LL}$, which integrates contextual information from both visible and thermal features.

\noindent\textbf{Frequency Fusion and Reconstruction.}
The concatenated high-frequency components from the visible modality, $\bm{F}^v_{{LH}}$ and $\bm{F}^v_{{HL}}$, are used to enhance the fused low-frequency representation through element-wise multiplication and addition:

\begin{equation}
\bm{F}'_{LL} = \bm{F}_{LL} \cdot [\bm{F}^v_{{LH}}, \bm{F}^v_{{HL}}] + \bm{F}_{LL}, 
\end{equation}

Finally, the processed low-frequency component $\bm{F}'_{LL}$, along with the high-frequency components $\bm{F}^v_{LH}$, $\bm{F}^v_{HL}$, and $\bm{F}^v_{HH}$, are passed through the Inverse DWT (IDWT) to reconstruct the final fused feature map:

\begin{equation} 
\bm{F}'_{LL}, \bm{F}^v_{{LH}}, \bm{F}^v_{{HL}}, \bm{F}^v_{{HH}} \xrightarrow{\text{IDWT}} \bm{F}'_{v_i}, 
\end{equation}

This fusion effectively combines semantic richness with spatial precision, enhancing detection performance for tiny objects in complex environments. Conventional optical-flow/correlation search and deformable convolution align and fuse features directly in the image domain, where semantic (low-frequency) and edge (high-frequency) information are inseparably mixed. By contrast, our wavelet-based CLFM first decomposes features into LL/LH/HL/HH sub-bands, aligns only the global LL context once, and then selectively gates high-frequency edges. This band-wise strategy reduces the alignment cost to $mathcal{O}(HW)$, suppresses modality-specific noise, and better preserves the sharp edges that are crucial for tiny-object detection.

\subsection{Dynamic Alignment and Scale Refinement Module}
The Dynamic Alignment and Scale Refinement (DASR) module is designed to address spatial misalignment and scale inconsistencies between visible and thermal features, particularly in challenging environments like drone-based scenarios. As shown in Fig.~\ref{fig:architecture} (c), DASR comprises two main components: the Adaptive Alignment Module (AAM) for pixel-level alignment and the Dynamic Scale Refinement (DSR) mechanism for multi-scale feature adjustment.

\noindent\textbf{Adaptive Alignment Module.}
The AAM focuses on aligning the visible feature map $\bm{F}_{v_i}$ with the thermal feature map $\bm{F}_{t_i}$ at the pixel level by learning spatial offsets. Specifically, the AAM predicts spatial offsets for each pixel in the visible feature map using a series of convolutional layers, where the thermal feature map serves as the reference. The input to this offset branch is the channel-wise concatenation of $\bm F_{v_i}$ and $\bm F_{t_i}$; the branch is supervised implicitly, its parameters are updated end-to-end by the task losses and, in particular, by the target-specific KL divergence that penalizes cross-modal mismatch. No extra pixel-level labels are required. The learned offsets are then applied through a grid sampling operation to adjust the spatial positioning of the visible features:

\begin{align}
{\Delta} \bm{V}_\mathit{i} &= \mathrm{tanh}(\mathrm{norm}(\mathrm{conv}_\mathit{k_i}(\bm{F}_{{v}_\mathit{i}}))),
\end{align}

where $\bm{F}_{v_\mathit{i}}$ is the visible feature map at stage $i$, and ${\Delta} \bm{V}_\mathit{i}$ represents the predicted offset map. The offsets are then used in the grid sampling process to adjust the visible feature map:

\begin{align}
\bm{F}'_{{v}_\mathit{i}}(x, y) &= \bm{\mathcal{G}}(\bm{F}_{{v}_\mathit{i}}, x + {\Delta} \bm{V}_\mathit{i}^\mathit{x}(x, y), y + {\Delta} \bm{V}_\mathit{i}^\mathit{y}(x, y)),
\end{align}
\begin{align}
\bm{F}'_{{t}_\mathit{i}}(x, y) &= \bm{\mathcal{G}}(\bm{F}_{{t}_\mathit{i}}, x, y),
\end{align}

where $\bm{\mathcal{G}}(\cdot)$ denotes the grid sampling operation, and $\bm{F}'_{{v}_\mathit{i}}$ and $\bm{F}'_{{t}_\mathit{i}}$ are the aligned feature maps. This pixel-level alignment ensures effective cross-modal feature fusion by minimizing spatial discrepancies.

\noindent\textbf{Dynamic Scale Refinement.}
Once the features are aligned, the DSR mechanism further refines them by addressing scale inconsistencies. The DSR mechanism employs convolution kernels of varying sizes, including standard $3 \times 3$ convolutions along with dilated convolutions. The output feature maps from these different convolutions are dynamically weighted and fused, allowing the module to emphasize the most relevant features across different scales. Additionally, Global Average Pooling (GAP) is applied to capture global context, followed by a $1 \times 1$ convolution to reduce dimensionality and enhance feature representation. The final refined feature map combines the outputs from the different convolutions and GAP to capture both local details and broader context effectively.

The integration of AAM and DSR ensures that DASR not only aligns features at the pixel level but also refines them continuously across multiple scales, thereby improving cross-modal feature fusion. Unlike the discrete-hypothesis scale search adopted in the Scale-and-State-Awareness (SSA) tracker~\cite{qi2019robust}, which evaluates a small set of fixed rescaling factors and selects the best one via correlation-filter response. DSR employs a single, learnable multi-branch convolution block whose branch weights are modulated by a global context gate derived from concatenated RGB-T features. This design offers two key advantages: (1) continuous scale adaptation with $\mathcal{O}(HW)$ complexity instead of the $\mathcal{O}(kHW)$ overhead incurred by enumerating k scale candidates, and (2) cross-modal scale reasoning, because the gating vector is conditioned jointly on visible and thermal cues. Together with pixel-level alignment from AAM, this dual mechanism is particularly effective for tiny-object detection under occlusion, variable lighting, and cluttered backgrounds, leading to more accurate localization and greater robustness.

\subsection{GeoShape-Based Label Assignment}

To obtain stable positives for tiny objects, we design a label-assignment rule that combines a dual-calculation scheme with a GeoShape similarity metric. GeoShape extends plain IoU by jointly modeling three normalized cues—center distance, aspect-ratio difference, and IoU—thereby reducing the sensitivity of classical IoU assignment to tiny shifts.

\noindent\textbf{Dual-calculation rule.}
Given an anchor box $\bm{b}$, a predicted box $\bm{p}$, and the ground truth box $\bm{g}$, we compute GeoShape similarity $\bm{\psi}(\cdot, \bm{g})$ for both $\bm{b}$ and $\bm{p}$ and keep the larger one:

\begin{equation}
s = \max(\bm{\psi}(\bm{b}, \bm{g}), \bm{\psi}(\bm{p}, \bm{g})),
\end{equation}

ensuring that a sample is retained as long as either its anchor or its current prediction matches the ground truth well. This two-way check makes the assignment robust throughout training.

\noindent\textbf{GeoShape similarity metric.} For an anchor or prediction box $\bm{a}=(c_x,c_y,w,h)$ and the ground truth box $\bm{g}=(c'_x,c'_y,w',h')$, we first compute:

\begin{equation}
d_c = \sqrt{\frac{(\mathit{c}_{x} - \mathit c'_{x})^2}{(\mathit{w} + \mathit{w'})^2} + \frac{(\mathit{c}_{y} - \mathit c'_{y})^2}{(\mathit{h} + \mathit h')^2}}, d_r=\left|\log(\frac{w}{h})-\log(\frac{w'}{h'})\right|,
\end{equation}

where $d_c$ measures center misalignment and $d_r$ captures aspect-ratio mismatch; both are scale-invariant by construction. Let IoU$(\bm{a}, \bm{g})$ be their overlap. The final similarity is:

\begin{equation}
\bm{\psi}(\bm{a}, \bm{g})=\exp \left[ -(d_c+\gamma \cdot d_r+\beta \cdot (1-\text{IoU}(\bm{a}, \bm{g})))\right],
\end{equation}

with fixed weights $\gamma=2$ and $\beta=1$. Compared with RFLA \cite{xu2022rfla} and SIWD \cite{zhang2023drone}, which employ two-dimensional Gaussian-based distributions focused primarily on localization accuracy, GeoShape explicitly integrates three geometric cues—center distance, aspect-ratio difference, and IoU—into a unified similarity metric without additional temperature hyper-parameters.

By fusing position, scale, and shape information in a scale-normalized form, the GeoShape metric tolerates small localization noise yet penalizes boxes that are geometrically inconsistent with the target. Coupled with the dual-calculation rule, it yields a label set that is both adaptive and reliable, which is crucial for tiny object detection in cluttered or occluded scenes.

\begin{table*}[t]
\centering
\renewcommand{\arraystretch}{1.4}
\footnotesize
\caption{Performance comparison with state-of-the-art methods on the RGBTDronePerson dataset~\cite{zhang2023drone}. All models, including ours, utilize a ResNet-50 backbone with an FPN. Best results are highlighted in bold.}
\resizebox{\textwidth}{!}
{%
\begin{tabular}{l|ccccccc|ccc}
\toprule
{Method} & {mAP$_\text{25}^{\text{all}}$} & {mAP$_\text{50}^{\text{tiny}}$} & {mAP$_\text{50}^\text{tiny1}$} & {mAP$_\text{50}^\text{tiny2}$} & {mAP$_\text{50}^\text{tiny3}$} & {mAP$_\text{50}^\text{small}$} & {mAP$_\text{50}^{\text{all}}$} & {Flops (G)} & {Params (M)} & {FPS} \\ 
\midrule
Cascade R-CNN†~\cite{cai2018cascade} & 
42.47 & 31.99 & 0.00 & 29.43 & 37.77 & 33.61 & 31.55 & 76.23 & 92.35 & 11.8 \\
RetinaNet†~\cite{lin2017focal}       & 
38.92 & 23.34 & 4.69 & 13.57 & 32.66 & 15.77 & 22.87 & 49.48 & 59.56 & \textbf{19.4} \\
FCOS†~\cite{tian2020fcos}            & 
45.21 & 30.71 & 9.40 & 22.73 & 34.87 & 26.00 & 29.89 & \textbf{47.21} & 78.32 & 19.2 \\
ATSS†~\cite{zhang2020bridging}       & 53.14 & 37.47 & 16.92 & 24.16 & 43.59 & 24.62 & 36.24 & 48.73 & \textbf{55.31} & 19.0 \\
GFL†~\cite{li2020generalized}        & 56.91 & 41.67 & 12.47 & 30.23 & 47.68 & 26.72 & 39.74 & 49.22 & 55.45 & 18.7 \\
FCOS† w/ RFLA~\cite{xu2022rfla}      & 51.41 & 39.45 & 25.87 & 30.93 & 44.31 & 25.32 & 38.20 & 110.62 & 78.32 & 13.9 \\
QueryDet†~\cite{yang2022querydet}    & 55.16 & 37.75 & 17.89 & 24.90 & 44.05 & 25.52 & 37.07 & 121.67 & 62.58 & 12.8 \\
TINet~\cite{luo2018tinet} & 40.34 & 28.60 & 0.00 & 24.12 & 34.99 & \textbf{34.97} & 28.30 & 92.80 & 100.86 & 13.0 \\
CFT~\cite{qingyun2021cross} & 37.32 & 22.83 & 16.72 & 18.14 & 27.52 & 8.14 & 22.69 & 112.02 & 206.26 & 10.8 \\
HRFuser~\cite{broedermann2023hrfuser} & 33.24 & 22.50 & 0.00 & 26.73 & 26.26 & 23.85 & 22.23 & 54.17 & 55.47 & 4.5 \\
QFDet~\cite{zhang2023drone} & 57.34 & 44.04 & 20.27 & 30.09 & 50.36 & 26.78 & 42.08 & 81.43 & 60.19 & 14.2 \\
QFDet*~\cite{zhang2023drone} & 61.62 & 48.75 & 22.15 & 37.91 & 53.71 & 28.41 & 46.72 & 242.82 & {60.25} & 5.7 \\
\midrule
\textbf{COXNet (Ours)} & 59.01 & 47.18 & \textbf{27.37} & 35.55 & 52.56 & 29.74 & 45.57 & 51.27 & 71.11 & 17.6 \\ 
\textbf{COXNet* (Ours)} & \textbf{62.76} & \textbf{51.82} & 23.08 & \textbf{40.10} & \textbf{56.76} & {30.89} & \textbf{50.04} & 123.59 & 75.86 & 12.9 \\ 
\bottomrule
\end{tabular}%
}
\begin{tablenotes}
\footnotesize
\item[*] † indicates methods adapted for the RGBT baseline detector. * denotes models utilizing detection heads with P2-P6 feature maps. 
\end{tablenotes}
\label{tab:performance_comparison}
\end{table*}

\begin{figure*}[t]
    \centering
    \includegraphics[width=1.0\linewidth]{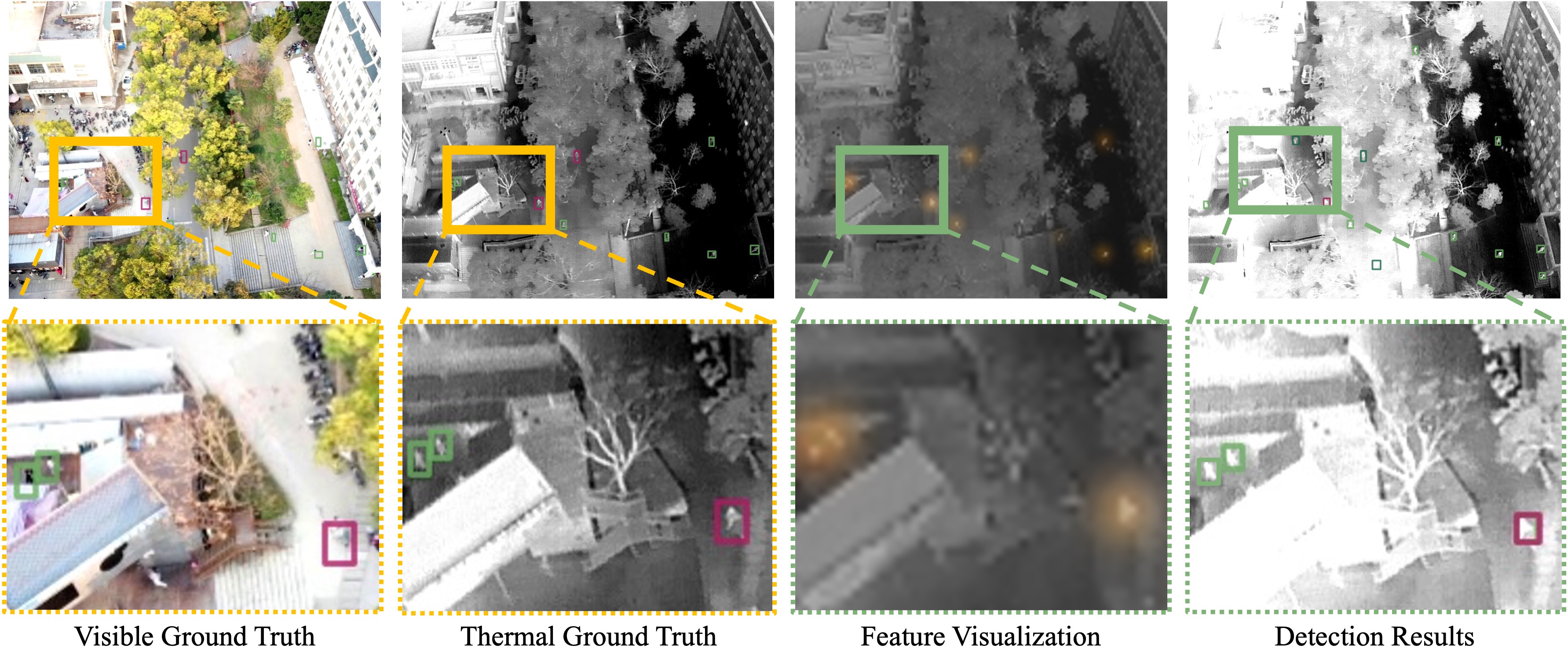}
    \caption{Examples of feature maps on the RGBTDronePerson dataset. The left column shows the original visible and thermal ground truth images, while the right column presents zoomed-in views of feature visualizations and final detection results. The bounding boxes highlight detected persons and riders. \textcolor{darkgreen}{Green} boxes indicate persons, while \textcolor{darkred}{red} boxes denote riders. \textcolor{orange}{Orange} regions denote strong feature response.}
    \label{fig:feature_droneperson}
\end{figure*}

\subsection{Target-Specific Cross-Modal Consistency Loss}
To stabilise training under modality mis-alignment, COXNet augments the
standard Generalised Focal Loss (GFL) formulation~\cite{li2020generalized} with an object-centered Kullback–Leibler term. The total objective is:

\begin{equation}
L_{\text{total}}
   = L_{\text{cls}} + L_{\text{reg}} + L_{\text{ctr}} + \lambda\cdot L_{\text{KL}},
\label{eq:loss_total}
\end{equation}

where $L_{\text{cls}}$, $L_{\text{reg}}$ and $L_{\text{ctr}}$ are the classification, regression and centerness losses in GFL~\cite{li2020generalized}, and $L_{\text{KL}}$ is the proposed cross-modal regularizer.

For every ground-truth box we enlarge the box by a factor of~1.5 and crop the corresponding tensor regions from the fused feature map and the thermal feature map. After softmax normalization along the channel axis, each region is viewed as a categorical distribution over feature channels.

Inside every enlarged region we treat the channel-wise softmax responses of the fused feature map and the thermal feature map as two probability distributions and minimize their bidirectional KL divergence. This object-centered penalty supplies dense gradients that encourage the fused feature map to retain sharp thermal cues and drive the thermal feature map towards the fused representation, without affecting background pixels.

\section{Experiments}

\subsection{Implementation Details}

All experiments were implemented using PyTorch and the mmdetection toolbox~\cite{mmdetection}. Input images were resized to a resolution of $640 \times 512$ across all datasets for consistency. We employed the Generalized Focal Loss (GFL) framework~\cite{li2020generalized}, utilizing a ResNet-50 backbone pre-trained on ImageNet, balancing computational efficiency with detection accuracy for tiny objects under diverse conditions.
Training was conducted using Stochastic Gradient Descent (SGD) with a momentum of 0.9 and a weight decay of 0.0001 for 12 epochs. An initial learning rate of 0.01 was reduced by a factor of 10 once validation performance plateaued. All experiments were performed on a single NVIDIA 2080Ti GPU.

To comprehensively evaluate COXNet's performance across various object scales, we categorized the targets into four sizes, following the COCO dataset protocol~\cite{lin2014microsoft}. 

\subsection{Results on RGBTDronePerson}
\noindent\textbf{Data and Experimental Setup.}
The experiments were conducted on the RGBTDronePerson dataset, which presents significant challenges due to its high proportion of tiny objects, modality imbalance, and spatial misalignment. We evaluated COXNet's performance against state-of-the-art methods, following the evaluation protocol used in the RGBTDronePerson dataset~\cite{zhang2023drone}.


\begin{figure*}[t]
\centering
\includegraphics[width=1.0\textwidth]{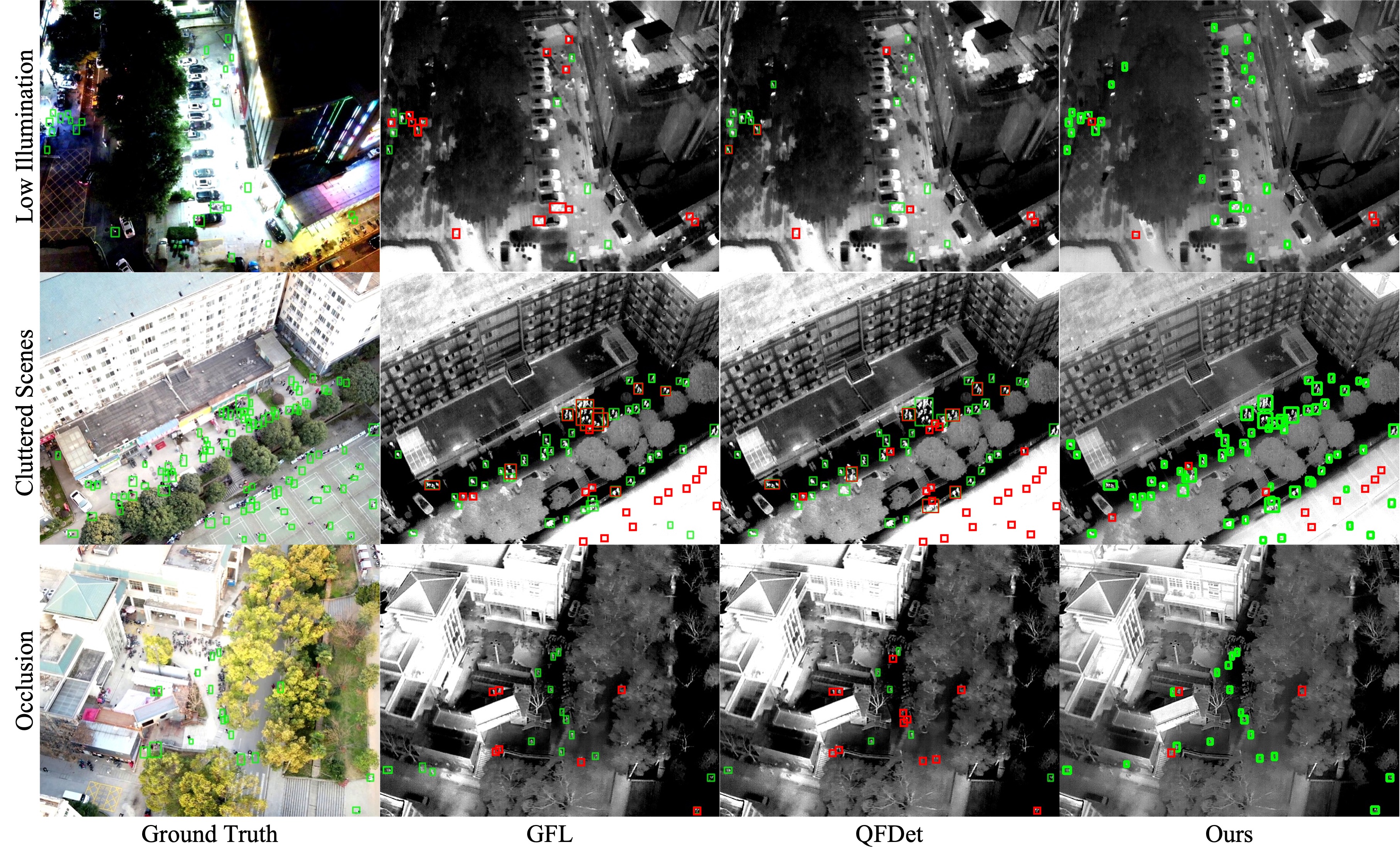}
\caption{Qualitative results on the RGBTDronePerson dataset. COXNet outperforms GFL and QFDet, particularly in detecting tiny, occluded objects. \textcolor{darkgreen}{Green} boxes indicate detected objects, while \textcolor{darkred}{red} boxes denote missed detections.}
\label{fig:qualitative_rgbtdroneperson}
\end{figure*}

\noindent\textbf{Main Results.}
COXNet* significantly outperforms all state-of-the-art methods on the RGBTDronePerson dataset, achieving a mAP$_{50}$ of 50.04\% and a mAP$_{25}$ of 62.76\%, as presented in Table~\ref{tab:performance_comparison}. These results highlight COXNet's superiority in detecting tiny and occluded objects, particularly in challenging conditions, with notable gains in mAP$_{50}^{\text{tiny}}$ (51.82\%), outperforming previous leading models like QFDet. The consistent improvement across metrics such as mAP$_{50}^\text{tiny2}$ and mAP$_{50}^\text{tiny3}$, where COXNet* achieves 40.10\% and 56.76\%, respectively, underlines its robustness in handling diverse and small-scale object detection tasks.

Furthermore, despite achieving the highest detection performance, COXNet maintains competitive efficiency with a moderate computational cost of 51.27 GFLOPs. Compared to methods like QFDet*, which require 242.82 GFLOPs, COXNet strikes a superior balance between accuracy and computational demand, making it suitable for real-time deployment in UAV-based and edge scenarios.

These comprehensive results emphasize the effectiveness of our architectural innovations, including the DASR and CLFM modules, in enhancing detection capabilities while balancing resource requirements, positioning COXNet as a state-of-the-art solution for RGBT tiny object detection.


\noindent\textbf{Visualization of Feature Maps and Analysis}  
To further understand COXNet's effectiveness, we provide feature map visualizations, as shown in Fig.~\ref{fig:feature_droneperson}. These examples illustrate the improved alignment and fusion of multi-modal features, which enhances the detection of small objects. The visualizations demonstrate how COXNet successfully integrates visible and thermal information, resulting in superior detection performance for challenging targets such as tiny and occluded objects.

The visualizations of the attention maps further reveal COXNet's ability to focus on regions of high significance, effectively reducing the influence of background noise. This focus contributes to enhanced detection accuracy, especially for small targets, by ensuring that crucial features are prioritized during processing.

Additionally, an analysis of feature distributions before and after fusion highlights the role of cross-modal feature alignment in improving detection quality. We also analyzed the training dynamics through a study of the loss landscape, which indicated that our design facilitates smoother gradients and more stable convergence, reinforcing COXNet's robustness in challenging detection tasks.

\noindent\textbf{Efficiency.}
Balancing detection accuracy and inference speed is critical for real-time applications, especially in resource-constrained environments like drone-based surveillance. As illustrated in Table~\ref{tab:performance_comparison}, COXNet* achieves an inference speed of 12.9 FPS, while delivering the highest detection accuracy with an mAP$_{50}$ of 50.04\%. Despite the added complexity of the DASR module and the detailed cross-layer feature fusion, COXNet maintains competitive speed, outperforming models like QFDet and QueryDet in both accuracy and computational efficiency.

The efficiency of COXNet is reflected not only in its higher mAP$_{50}$ values but also in the balance it achieves between speed and computational complexity, represented by the FLOPs. By optimizing the design to selectively leverage detailed feature fusion without overwhelming computational requirements, COXNet emerges as an optimal solution for real-time detection tasks in complex RGBT environments, where both accuracy and efficiency are paramount.

\noindent\textbf{Qualitative Results.}
Fig. \ref{fig:qualitative_rgbtdroneperson} offers a visual counterpart to the numerical gains reported earlier. Across low-illumination shots, crowded streets, and severely cluttered backgrounds, COXNet recovers many tiny or partly occluded pedestrians that GFL and QFDet fail to register. In dim light, the network pinpoints targets whose thermal signatures are faint and whose RGB contrast is poor; in dense urban frames, it suppresses background activations and cleanly separates overlapping instances, resulting in noticeably fewer false negatives and broader true-positive coverage. These improvements arise from the complementary effects of the CLFM’s band-wise fusion, the DASR’s pixel- and scale-level alignment, and the GeoShape label strategy, which together deliver sharper localisation and greater recall.

Residual errors are now dominated by two classic hard cases. First, pedestrians rendered in fewer than five pixels remain difficult: at that scale the signal-to-noise ratio in both modalities is simply too low for reliable discrimination. Second, extreme partial occlusion—for example a torso hidden behind dense foliage or street furniture—can still obscure the most informative features.

\begin{figure*}[ht]
\centering
\includegraphics[width=1.0\textwidth]{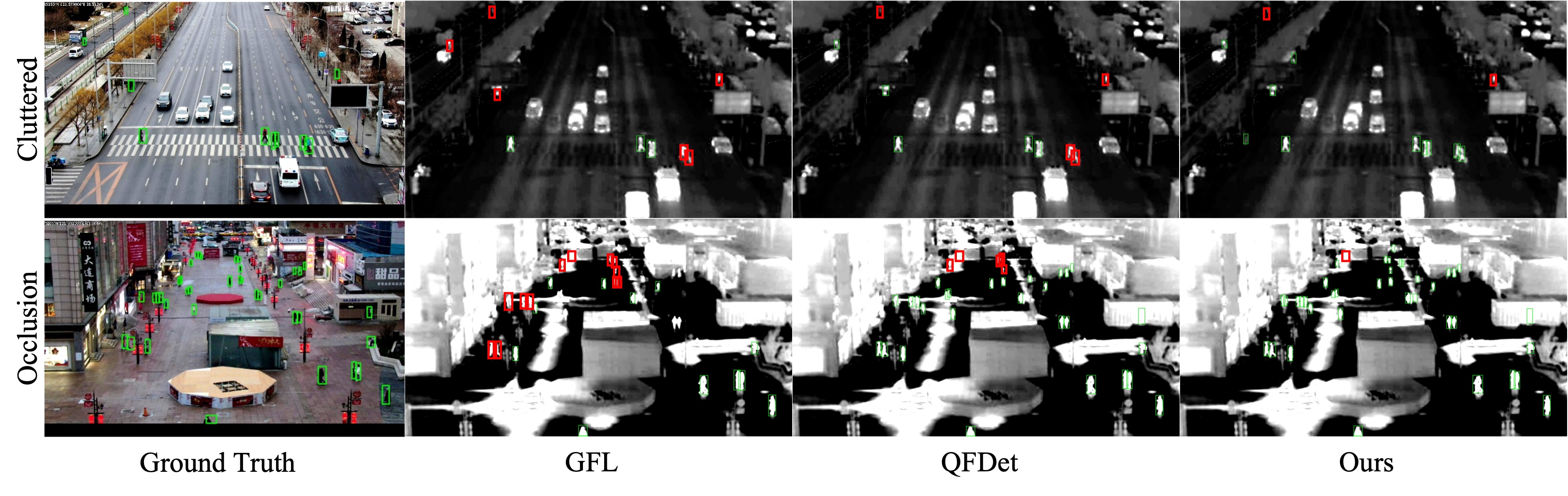}
\caption{Qualitative results on the VTUAV-det dataset under challenging conditions, including cluttered scenes and occlusion. \textcolor{darkgreen}{Green} boxes indicate detected objects, while \textcolor{darkred}{red} boxes denote missed detections.}
\label{fig:qualitative_vtuav}
\end{figure*}

\subsection{Results on VTUAV-det}
\noindent\textbf{Data and Experimental Setups.}
The VTUAV-det dataset~\cite{zhang2023drone} is a benchmark for evaluating tiny object detection in drone-based aerial imagery, containing 11,392 training and 5,378 testing RGBT image pairs across three categories: vehicle, person, and cyclist. It presents typical challenges such as occlusion, varying illumination, and numerous small objects due to the high-altitude perspective, making it suitable for evaluating the effectiveness of cross-modal fusion techniques.

COXNet was evaluated on VTUAV-det to assess its ability to address these challenges, particularly for small and tiny objects, through precise cross-modal alignment and improved fusion.

\noindent\textbf{Main Results.}
COXNet* demonstrates strong performance on the VTUAV-det dataset, achieving a mAP$_{50}$ of 76.1\% and outperforming all state-of-the-art methods, as shown in Table~\ref{tab:vtuav_performance_comparison}. Specifically, for small object detection, COXNet* achieves a mAP$_s$ of 18.6\%, surpassing all other methods and showcasing its ability to handle small targets effectively, even under challenging aerial conditions with occlusion and limited pixel representation.

The improvements in small object detection can be attributed to COXNet's Dynamic Scale Refinement (DSR) and Adaptive Alignment Module (AAM), which are designed to enhance feature extraction and alignment. COXNet* also performs competitively for medium and large objects, with mAP$_m$ of 32.6\% and mAP$_l$ of 56.8\%. The focus on small object detection may result in slightly less pronounced improvements for larger objects, where more global context is typically required.

Furthermore, COXNet* achieves a mAP$_{75}$ of 25.1\%, demonstrating its robustness under stricter localization requirements, along with an inference speed of 15.0 FPS, providing a good balance between detection accuracy and real-time performance.

\begin{table}[ht]
\centering
\renewcommand{\arraystretch}{1.4}
\footnotesize
\setlength{\tabcolsep}{3.7pt}
\caption{Performance comparison with state-of-the-art methods on VTUAV-det~\cite{zhang2023drone}. The best results are highlighted in bold.}
{%
\begin{tabular}{l|cccccccc}
\toprule
{Method} & mAP & {mAP$_\text{50}$} & {mAP$_\text{75}$} & {mAP$_\text{s}$} & {mAP$_\text{m}$} & {mAP$_\text{l}$} & FPS \\ 
\midrule
ATSS†~\cite{zhang2020bridging} & 21.4 & 52.7 & 13.9 & 5.9 & 20.8 & 45.2 & \textbf{25.1} \\
GFL†~\cite{li2020generalized} & 29.8 & 67.8 & 22.2 & 10.3 & 29.7 & 55.7 & 23.6 \\
QueryDet†~\cite{yang2022querydet} & 29.5 & 68.9 & 20.2 & 7.8 & 29.9 & 53.5 & 14.6\\
CFT~\cite{qingyun2021cross} &8.7 & 29.3 & 2.4 & 3.9 & 8.5 &23.4 & 8.3\\
HRFuser~\cite{broedermann2023hrfuser} & 25.9 & 55.9&20.1&2.7&27.9 &51.9&5.6\\
TINet~\cite{luo2018tinet}&26.8&59.4&20.1&1.2&29.0&53.7&14.5\\
QFDet~\cite{zhang2023drone} & 31.1 & 70.4 & 22.9 & 12.5 & 20.4 & 56.8 & 15.3 \\
QFDet*~\cite{zhang2023drone} & 33.3 & 75.5 & 24.2 & 18.1 & 32.4 & \textbf{57.2} & 9.4 \\
\midrule
\textbf{COXNet(Ours)} & 31.5 & 71.8 & 23.1 & 15.3 & 30.6 & 56.0 & 21.2 \\
\textbf{COXNet*(Ours)} & \textbf{33.5}  & \textbf{76.1}  & \textbf{25.1} & \textbf{18.6} & \textbf{32.6}   & 56.8 & 15.0 \\ 
\bottomrule
\end{tabular}%
}
\begin{tablenotes}
\footnotesize
\item[*] † denotes methods modified for the RGBT baseline detector. * indicates that the detection head utilizes P2-P6 feature maps.
\end{tablenotes}
\label{tab:vtuav_performance_comparison}
\end{table}

\noindent\textbf{Qualitative Results.}
Qualitative results on the VTUAV-det dataset further illustrate COXNet's ability to detect tiny and occluded objects in various challenging real-world conditions, including cluttered backgrounds and complex aerial perspectives. As shown in Fig.~\ref{fig:qualitative_vtuav}, COXNet consistently enhances object localization precision compared to baseline methods, demonstrating a marked improvement in detecting a higher number of small targets with minimal missed detections. This highlights its sensitivity and robustness in handling challenging detection tasks in cluttered environments.

\begin{figure*}[ht]
\centering
\includegraphics[width=1.0\textwidth]{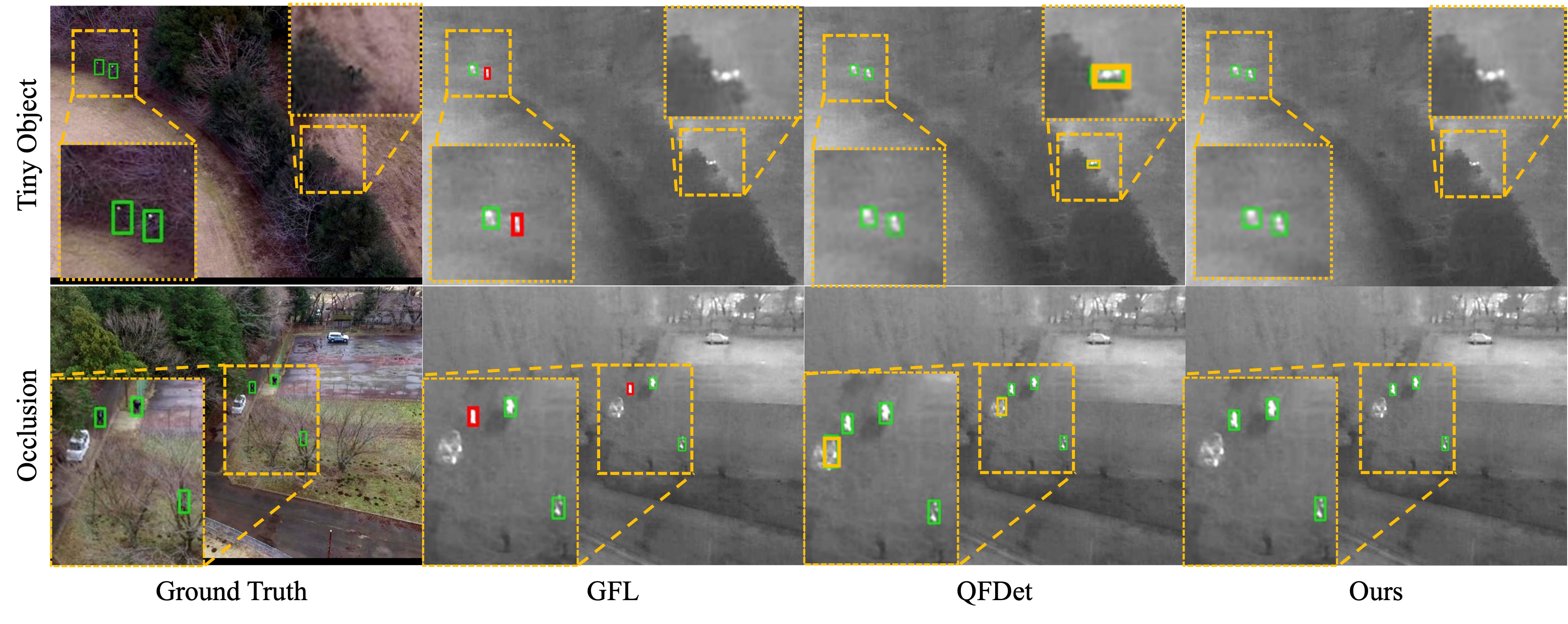}
\caption{Qualitative results on the NII-CU dataset under low illumination and occlusion conditions. COXNet demonstrates superior performance compared to baseline methods such as GFL and QFDet. \textcolor{darkgreen}{Green} boxes represent correct detections, \textcolor{darkred}{red} boxes indicate missed detections, and \textcolor{orange}{orange} boxes denote false detections.}
\label{fig:qualitative_niicu}
\end{figure*}

\subsection{Results on NII-CU}
\noindent\textbf{Data and Experimental Setups.}
The NII-CU dataset~\cite{speth2022deep} contains 6,000 pairs of RGBT images with 19,000 annotated instances across three categories: pedestrian, vehicle, and cyclist. It presents challenges related to urban clutter, adverse weather, and occlusion, making it a suitable benchmark for evaluating the robustness of detection models under complex aerial conditions.

COXNet was evaluated on the NII-CU dataset to demonstrate its effectiveness in addressing scale variability, perspective changes, and modality imbalances, particularly for tiny and occluded objects.

\noindent\textbf{Main Results.}
COXNet demonstrates superior performance on the NII-CU dataset, as shown in Table~\ref{tab:niicu_performance_comparison}. Specifically, COXNet achieves the highest mAP$_\text{50}$ of 98.2\%, outperforming all other methods. Additionally, COXNet* delivers a mAP of 65.4\% and a mAP$_\text{75}$ of 79.6\%, surpassing previous state-of-the-art results. Moreover, COXNet maintains a favorable balance between accuracy and efficiency, with an inference speed of 17.9 FPS, making it suitable for real-time applications, even in challenging, resource-constrained environments.

The strong performance of COXNet is largely attributed to its optimized Cross-Layer Fusion Module (CLFM). Unlike Transformer-based models like CFT~\cite{qingyun2021cross}, which rely on computationally intensive fusion strategies, CLFM is designed to address the unique challenges of RGBT data—such as modality imbalance and spatial misalignment—through efficient convolutional operations. This allows COXNet to balance high detection accuracy with computational efficiency, while ensuring precision in detecting small and occluded objects.

The results demonstrate the effectiveness of our CNN-based approach, particularly in scenarios requiring both high precision and real-time processing. COXNet's consistent performance in detecting small, occluded objects under varying lighting conditions further emphasizes its robustness.

\begin{table}[t]
\centering
\renewcommand{\arraystretch}{1.4}
\footnotesize
\setlength{\tabcolsep}{11.9pt}
\caption{Performance comparison with state-of-the-art methods on NII-CU~\cite{speth2022deep}. The bold font indicates the best results.}
{%
\begin{tabular}{l|ccccc}
\toprule
{Method} & mAP & {mAP$_\text{50}$} & {mAP$_\text{75}$} & {FPS}  \\ 
\midrule
ATSS†~\cite{zhang2020bridging} & 54.6 & 95.5 & 55.0 & \textbf{24.1} \\
GFL†~\cite{li2020generalized} & 61.0 & 96.7 & 71.2 & 19.6 \\
CFT~\cite{qingyun2021cross} & 51.2 & 95.1 & 58.7 & 9.2 \\
QFDet~\cite{zhang2023drone} & 58.3 & 96.7 & 65.3 & 17.3 \\
QFDet*~\cite{zhang2023drone} & 63.7 & 97.6 & 76.4 & 10.3 \\
\midrule
COXNet(Ours) & 61.4 & \textbf{98.2} & 70.5 & 17.9 \\
\textbf{COXNet*(Ours)} & \textbf{65.4} & 97.9 & \textbf{79.6} & 13.1 \\ 
\bottomrule
\end{tabular}%
}
\begin{tablenotes}
\footnotesize
\item[*] † denotes methods modified for the RGBT baseline detector. * indicates that the detection head utilizes P2-P6 feature maps.
\end{tablenotes}
\label{tab:niicu_performance_comparison}
\end{table}

\noindent\textbf{Qualitative Results.}
The qualitative results on the NII-CU dataset, shown in Fig.~\ref{fig:qualitative_niicu}, demonstrate that COXNet detects objects with greater accuracy and exhibits fewer missed and false detections compared to baseline methods like GFL and QFDet. Its superior performance is particularly evident in challenging scenarios, such as detecting tiny and occluded objects, highlighting its robustness for real-world multimodal object detection tasks.

\subsection{Ablation Studies}
\noindent\textbf{Effect of the Main Components.}
We conducted ablation experiments to evaluate the impact of the Dynamic Alignment and Scale Refinement (DASR) module, Cross-Layer Fusion Module (CLFM), and GeoShape similarity measure, as shown in Table~\ref{tab:ablation_study}. Each component significantly improved detection performance. Adding DASR increased mAP$_{50}^{\text{all}}$ from 39.90\% to 43.00\%, while GeoShape and CLFM improved it to 41.69\% and 44.08\%, respectively. The combination of DASR and GeoShape achieved an mAP$_{50}^{\text{all}}$ of 44.07\%, and integrating all components resulted in the best performance, reaching 46.23\%. These results demonstrate the complementary nature of the proposed modules, which together enhance detection accuracy, particularly for tiny objects.

\begin{table}[!t]
\centering
\footnotesize
\renewcommand{\arraystretch}{1.4}  
\caption{Ablation study of the main components of COXNet on the RGBTDronePerson dataset.}
\setlength{\tabcolsep}{2.3pt}
{%
\begin{tabular}{ccc|cccccccc}
\toprule
{DASR} & {GeoShape} & {CLFM} & {mAP$_{25}$} & {mAP$_{50}^{\text{tiny}}$} & {mAP$_{50}^{\text{small}}$} & {mAP$_{50}^{\text{all}}$} & Flops (G) \\ 
\midrule
$\times$ & $\times$ & $\times$ & 56.80 & 41.67 & 25.11 & 39.90 & \textbf{49.16} \\
\checkmark & $\times$ & $\times$ & 57.76 & 44.65 & 28.27 & 43.00 & 50.87 \\
$\times$ & \checkmark & $\times$ & 57.84 & 43.41 & 28.39 & 41.69 & 50.21 \\
$\times$ & $\times$ & \checkmark & 57.46 & 45.83 & 28.22 & 44.08 & 49.56 \\
\checkmark & \checkmark & $\times$ & 58.92 & 45.76 & 28.76 & 44.07 & 50.88 \\
\checkmark & $\times$ & \checkmark & 57.58 & 45.90 & \textbf{29.64} & 44.32 & 51.26 \\
$\times$ & \checkmark & \checkmark & 58.26 & 46.45 & 27.05 & 44.94 & 50.60 \\
\checkmark & \checkmark & \checkmark & \textbf{59.89} & \textbf{47.97} & 29.20 & \textbf{46.23} & 51.27 \\
\bottomrule
\end{tabular}%
\begin{tablenotes}
\footnotesize
\item[*] DASR: Dynamic Alignment and Scale Refinement Module. GeoShape: GeoShape-Based Label Assignment. CLFM: Cross-Layer Fusion Module.
\end{tablenotes}
}
\label{tab:ablation_study}
\end{table}

\begin{figure}[t]
    \centering
    \includegraphics[width=1.0\linewidth]{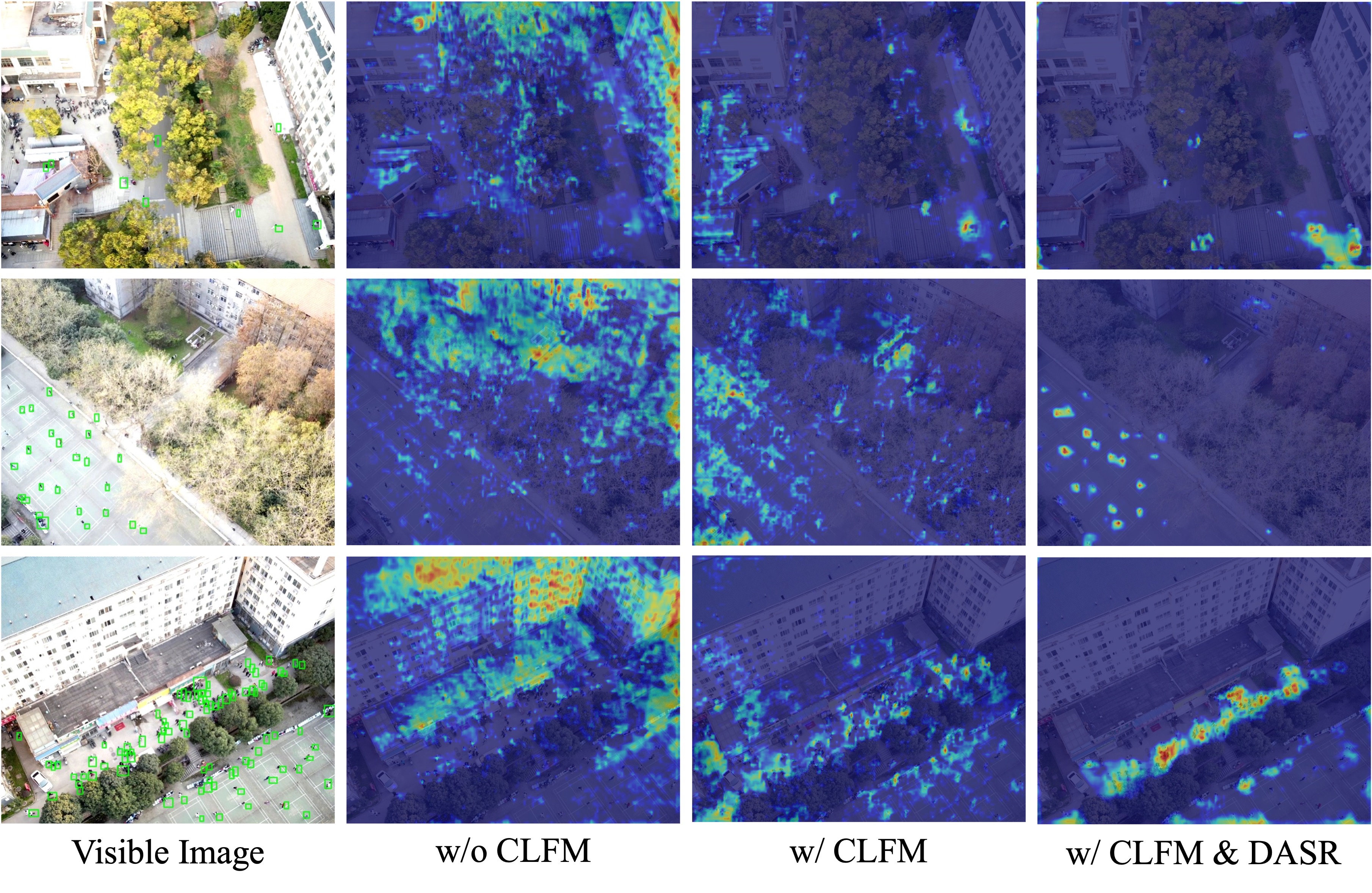}
    \caption{Feature map comparison on the RGBTDronePerson dataset. The first column shows visible reference images, while the second, third, and fourth columns present feature maps without CLFM, with CLFM, and with both CLFM and DASR.}
    \label{fig:ablation_components}
\end{figure}

\noindent\textbf{Impact of Cross-Layer Fusion Module.}
Table~\ref{tab:clfm_direct_comparison} examines four CLFM variants that incrementally activate its key components. Starting from the thermal-only baseline (row 1), simply injecting higher-level visible features lifts $\text{mAP}_{50}^{\text{all}}$ from 39.89\% to 43.92\% (row 2), confirming the benefit of cross-modal information even when the two streams are merged by Spatial Addition Fusion (SAF, i.e.\ element-wise addition in the spatial domain). Adding a lightweight deconvolution to match resolutions gives a further 1.31\% gain (row 3). When the full frequency-domain fusion—Discrete Wavelet Transform (DWT)—is enabled (row 4), performance reaches 46.23\%, yielding an overall 6.34\% improvement relative to the baseline and a 1.00\% margin over SAF with deconvolution. These results quantitatively verify that frequency-domain manipulation provides additional discriminative power beyond spatial-domain addition.

\begin{figure}[ht]
    \centering
    \includegraphics[width=1.0\linewidth]{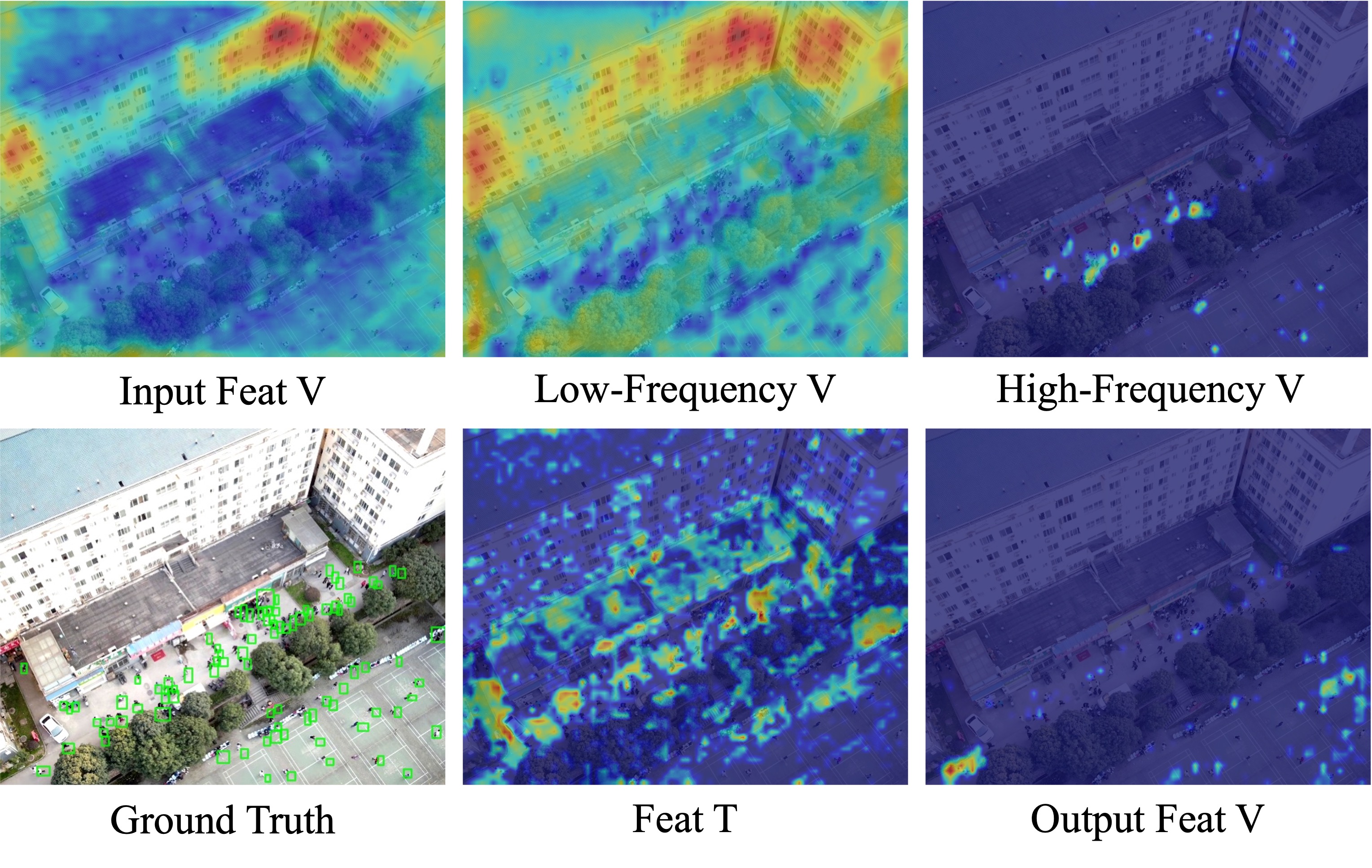}
    \caption{Visualization of CLFM wavelet decomposition. Low-frequency bands supply global context, high-frequency bands bring fine edges (and noise); gated fusion amplifies useful visible edges missed by thermal data, preparing features for later cross-modal fusion.}
    \label{fig:wavelet_vis}
\end{figure}

Fig.~\ref{fig:wavelet_vis} offers qualitative insight: the DWT cleanly separates low-frequency bands (global context) from high-frequency bands (fine edges).  CLFM first aligns and fuses the LL bands, then uses a learnable gate to re-inject visible-band edges while damping thermal noise. The resultant feature maps focus sharply on tiny targets that the thermal stream alone fails to highlight, furnishing a stronger input for the subsequent DASR module and ultimately boosting detection under challenging conditions.

\begin{table}[t]
\centering
\footnotesize
\renewcommand{\arraystretch}{1.4}  
\caption{Performance comparison of different CLFM configurations on the RGBTDronePerson dataset.}
\setlength{\tabcolsep}{2.5pt}  
{%
\begin{tabular}{ccc|cccccc}  
\toprule
\multicolumn{3}{c|}{{CLFM}} & \multirow{2}{*}{mAP$_{25}^{\text{all}}$} & \multirow{2}{*}{mAP$_{50}^{\text{tiny}}$} & \multirow{2}{*}{mAP$_{50}^{\text{small}}$} & \multirow{2}{*}{mAP$_{50}^{\text{all}}$} & \multirow{2}{*}{Flops (G)} \\  \cline{1-3}
{Higher-V} & {Deconv} & {DWT} \\
\midrule
$\times$ & $\times$ & $\times$ & 55.94 & 41.04 & 26.63 & 39.89 & 49.04 \\
\checkmark & $\times$ & $\times$ & 58.89 & 45.94 & 26.50 & 43.92 & \textbf{49.04} \\
\checkmark & \checkmark & $\times$ & 58.41 & 46.90 & 28.74 & 45.23 & 49.84 \\
\checkmark & \checkmark & \checkmark & \textbf{59.89} & \textbf{47.97} & \textbf{29.20} & \textbf{46.23} & 51.27 \\
\bottomrule
\end{tabular}%
}
\label{tab:clfm_direct_comparison}
\begin{tablenotes}
\footnotesize
\item[*] Higher-V: High-level visible features. Deconv: Upsampling deconvolution. DWT: Wavelet-based fusion ($\times$ denotes spatial addition).
\end{tablenotes}
\end{table}

\noindent\textbf{Impact of the DASR Module.}
We analyzed the effectiveness of the DASR module, which integrates the Adaptive Alignment Module (AAM) and Dynamic Scale Refinement (DSR), using the RGBTDronePerson dataset. As shown in Table~\ref{tab:om_msf_impact}, incorporating AAM alone improved mAP$_{50}^{\text{all}}$ from 41.73\% to 45.77\%, while DSR alone reached 41.73\%. Combining AAM and DSR yielded the best results, achieving an mAP$_{50}^{\text{all}}$ of 46.23\%, demonstrating the complementary nature of these components.

\begin{figure}[htbp]
    \centering
    \includegraphics[width=1.0\linewidth]{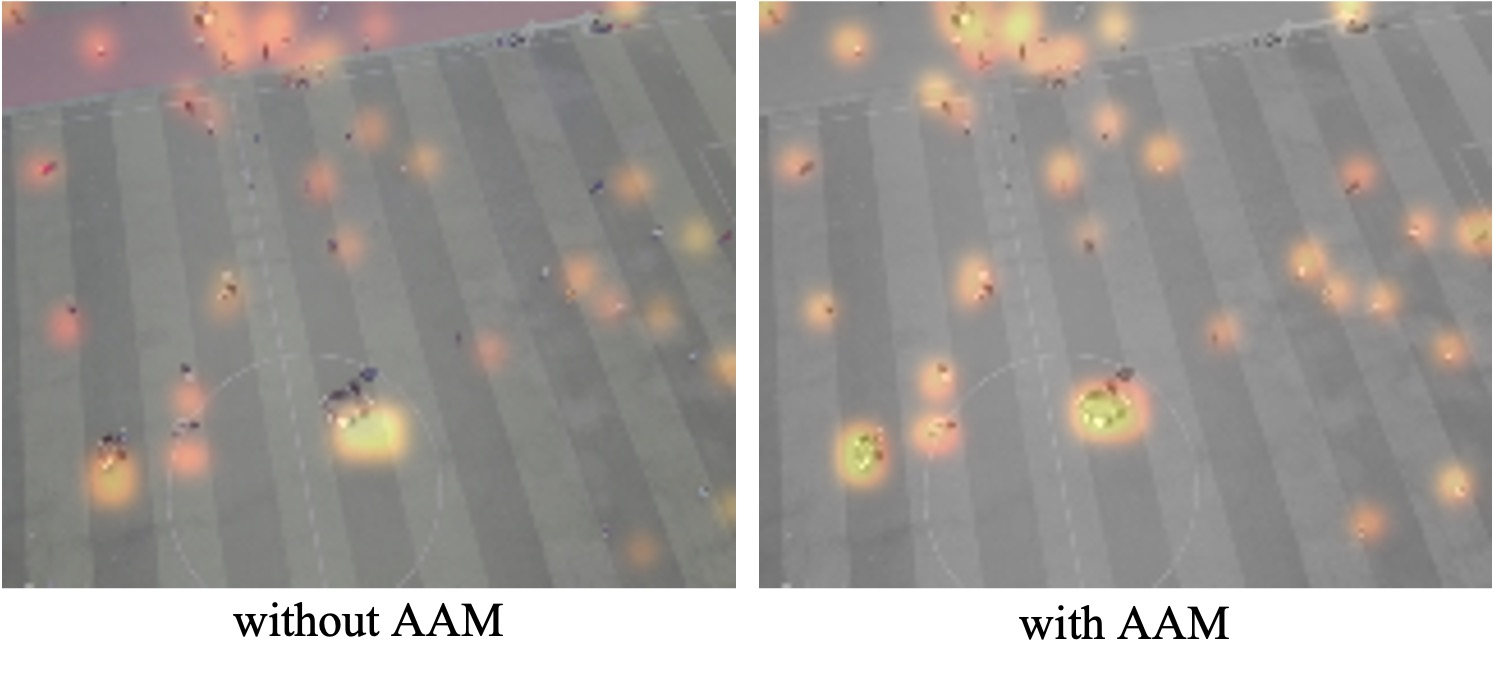}
    \caption{Feature map comparison with and without the Adaptive Alignment Module (AAM). The left image shows feature misalignment without AAM, while the right image demonstrates improved alignment achieved by AAM, resulting in enhanced object detection performance. \textcolor{orange}{Orange} regions denote strong feature response.}
    \label{fig:ablation_aam}
\end{figure}

We further evaluated alignment strategies within AAM, as summarized in Table~\ref{tab:aam_alignment_strategies}. The proposed adaptive offset-based alignment outperformed both Similarity Transformation (ST) and Fixed Grid Alignment (GA), achieving the highest detection accuracy. Feature map visualizations in Fig.~\ref{fig:ablation_aam} illustrate the superior alignment achieved by AAM, leading to enhanced feature quality and better detection performance.

\begin{table}[t]
\centering
\footnotesize
\renewcommand{\arraystretch}{1.4}  
\caption{Performance comparison of the Adaptive Alignment Module (AAM) and Dynamic Scale Refinement (DSR) in the DASR module on the RGBTDronePerson dataset.}
\setlength{\tabcolsep}{5.8pt}
{%
\begin{tabular}{cc|cccccc}
\toprule
\multicolumn{2}{c|}{{DASR}} & \multirow{2}{*}{mAP$_{25}$} & \multirow{2}{*}{mAP$_{50}^{\text{tiny}}$} & \multirow{2}{*}{mAP$_{50}^{\text{small}}$} & \multirow{2}{*}{mAP$_{50}^{\text{all}}$}  & \multirow{2}{*}{Flops (G)}\\  \cline{1-2}
{AAM} & {DSR}  & \\ 
\midrule
\checkmark & $\times$ & 59.36 & 47.21 & 28.66 & 45.77 & \textbf{50.07} \\
$\times$ & \checkmark & 57.15 & 43.19 & 29.02 & 41.73 & 50.76 \\
\checkmark & \checkmark & \textbf{59.89} & \textbf{47.97} & \textbf{29.20} & \textbf{46.23} & 51.27 \\ 
\bottomrule
\end{tabular}%
}
\label{tab:om_msf_impact}
\end{table}

\begin{table}[t]
\centering
\footnotesize
\renewcommand{\arraystretch}{1.4}  
\caption{Comparison of different alignment strategies in the Adaptive Alignment Module (AAM) on the RGBTDronePerson dataset.}
{%
\begin{tabular}{l|cccccc}
\toprule
{Strategy} & {mAP$_{25}$} & {mAP$_{50}^{\text{tiny}}$} & {mAP$_{50}^{\text{small}}$} & {mAP$_{50}^{\text{all}}$} & Flops (G)\\ 
\midrule
ST & 57.87 & 42.18 & 26.84 & 40.61 & 51.49 \\
Fixed GA & 58.44 & 46.75 & 29.13 & 45.10 & \textbf{51.21} \\
AAM & \textbf{59.89} & \textbf{47.97} & \textbf{29.20} & \textbf{46.23} & 51.27 \\ 
\bottomrule
\end{tabular}%
}
\label{tab:aam_alignment_strategies}
\begin{tablenotes}
\footnotesize
\item[*] ST: Similarity Transformation, Fixed GA: Fixed Grid Alignment.
\end{tablenotes}
\end{table}

\begin{table}[!h]
\centering
\footnotesize
\renewcommand{\arraystretch}{1.4}  
\caption{Ablation study on the impact of the KL divergence loss ($L_{\text{KL}}$) with different scaling factors ($\lambda$) on the RGBTDronePerson dataset.}
{%
\begin{tabular}{l|cccccc}
\toprule
$L_{\text{KL}}$ & {mAP$_{25}$} & {mAP$_{50}^{\text{tiny}}$} &{mAP$_{50}^{\text{small}}$} & {mAP$_{50}^{\text{all}}$} & Flops (G)\\ 
\midrule
None & 55.83 & 46.38 & 28.34 & 44.78 & 51.27 \\
\textbf{$\lambda$ = 0.1} & \textbf{59.89} & \textbf{47.97} & 29.20 & \textbf{46.23} & \textbf{51.27} \\ 
$\lambda$ = 0.5 & 59.24 & 46.74 & \textbf{31.56} & 45.37 & 51.27 \\
$\lambda$ = 1.0 & 58.46 & 46.69 & 27.35 & 44.79 & 51.27 \\
\bottomrule
\end{tabular}%
}
\label{tab:ablation_loss}
\end{table}

\begin{table}[H]
\centering
\footnotesize
\renewcommand{\arraystretch}{1.4}
\caption{Performance comparison of label assignment strategies (IoU, GIoU, SIWD, and GeoShape) on the RGBTDronePerson dataset.}
{%
\begin{tabular}{l|ccccccc}  
\toprule
{Strategy} & {mAP$_{25}$} & {mAP$_{50}^{\text{tiny}}$} &  {mAP$_{50}^{\text{small}}$} & {mAP$_{50}^{\text{all}}$} & Flops (G) \\ 
\midrule
IoU & 58.94 & 47.27 & 26.07 & 45.41 & 51.27 \\
GIoU & 58.28 & 46.16 & 27.98 & 44.33 & 51.27 \\
SIWD & 58.64 & 47.14 & 28.79 & 45.32 & 51.27 \\
\textbf{GeoShape} & \textbf{59.89} & \textbf{47.97} & \textbf{29.20} & \textbf{46.23} & \textbf{51.27} \\
\bottomrule
\end{tabular}%
}
\label{tab:geoshape_comparison}
\end{table}

\noindent\textbf{Effectiveness of the GeoShape Similarity Measure.} 
We evaluated the GeoShape similarity measure against IoU, GIoU, and SIWD~\cite{zhang2023drone} as label assignment strategies on the RGBTDronePerson dataset, as shown in Table~\ref{tab:geoshape_comparison}. GeoShape achieved the highest mAP$_{50}^{\text{tiny}}$ of 47.97\% and mAP$_{50}^{\text{all}}$ of 46.23\%, outperforming other methods. These results highlight GeoShape’s ability to better capture spatial and shape alignment, particularly for tiny objects, enhancing detection accuracy.

\noindent\textbf{Impact of KL Divergence Loss.}
We evaluated the impact of the KL divergence loss ($L_{\text{KL}}$) by varying its scaling factor ($\lambda$), as shown in Table~\ref{tab:ablation_loss}. The best performance was achieved with $\lambda = 0.1$, yielding an mAP$_{50}^{\text{all}}$ of 46.23\%. This demonstrates that $L_{\text{KL}}$ effectively improves cross-modal feature alignment, enhancing detection accuracy for tiny objects. Excessively high $\lambda$ values, such as 1.0, reduced performance, indicating the need for balanced contributions from $L_{\text{KL}}$.

\section{Conclusion}

We introduced COXNet, a framework for RGBT tiny object detection that addresses spatial misalignment and multi-modal fusion challenges. The Dynamic Alignment and Scale Refinement (DASR) module effectively corrects misalignments, while the Cross-Layer Fusion Module (CLFM) enhances feature integration, leading to improved detection in complex environments. Additionally, our GeoShape-based label assignment strategy refines small object localization, outperforming traditional IoU-based methods. Experiments on RGBTDronePerson, VTUAV-det, and NII-CU datasets demonstrate COXNet’s state-of-the-art performance and efficiency, making it suitable for real-time drone applications. Future work will extend COXNet to broader multi-modal tasks and enhance scalability.

{
\bibliographystyle{IEEEtran}
\bibliography{references}
}

\newpage
\begin{IEEEbiography}[{\includegraphics[width=1in,height=1.25in,clip,keepaspectratio]{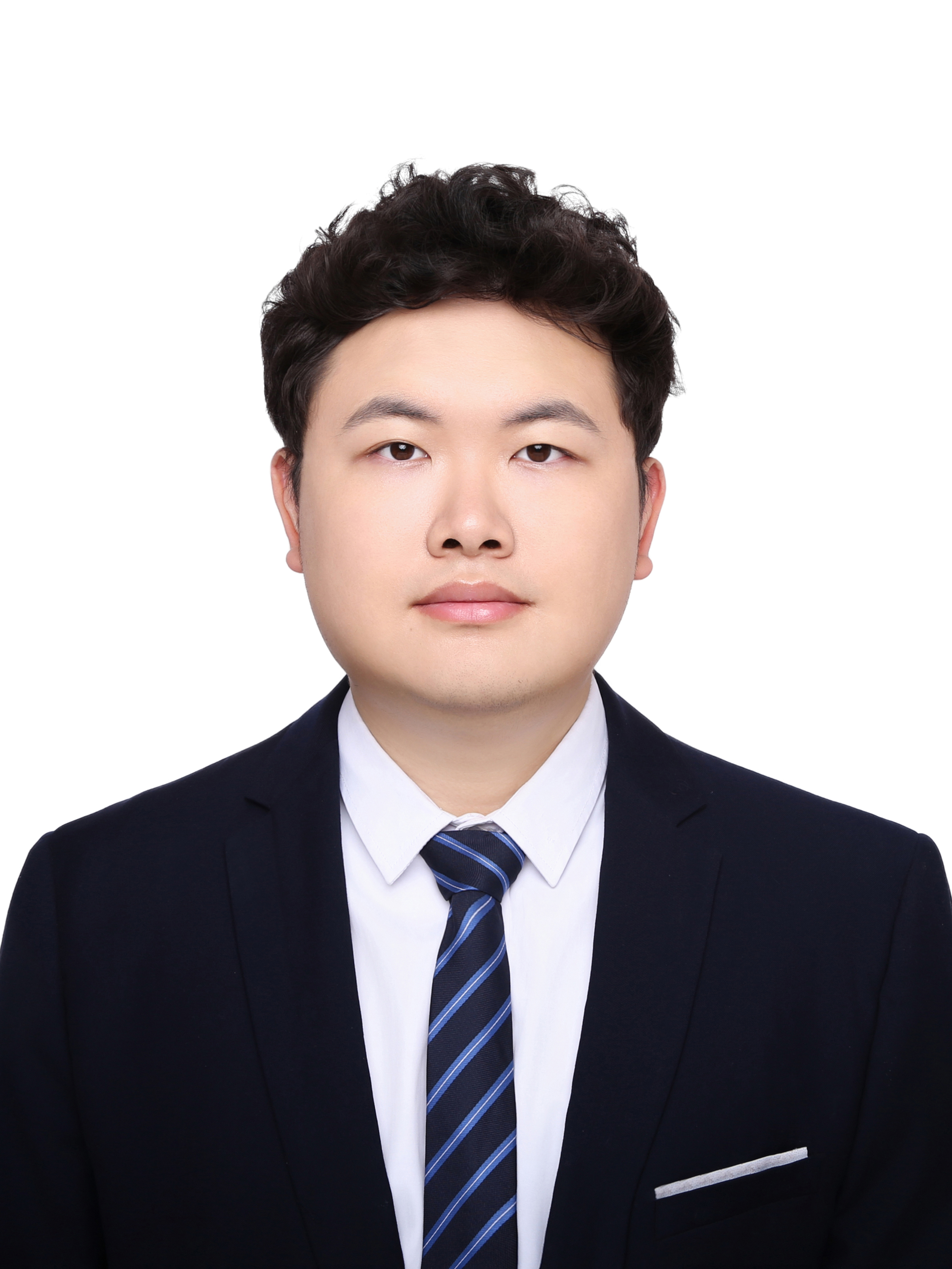}}]{Peiran Peng} received the master degree from the Institute of Microelectronics (IME) of the Chinese Academy of Sciences, Beijing, China, in 2021. He is currently a Ph.D. candidate with the School of Optoelectronics, Beijing Institute of Technology, Beijing, China. His research interests mainly include computer vision and real-time image/video processing.
\end{IEEEbiography}

\begin{IEEEbiography}[{\includegraphics[width=1in,height=1.25in,clip,keepaspectratio]{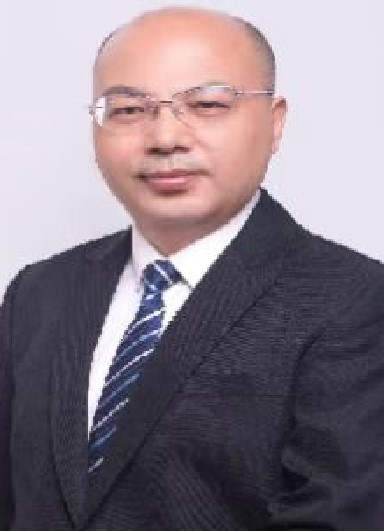}}]{Tingfa Xu} received the Ph.D. degree from the Changchun Institute of Optics, Fine Mechanics and Physics, Changchun, China, in 2004. He is currently a Professor with author'shool of Optics and Photonics, Beijing Institute of Technology, Beijing, China. His research interests include optoelectronic imaging and detection and hyperspectral remote sensing image processing.
\end{IEEEbiography}

\begin{IEEEbiography}[{\includegraphics[width=1in,height=1.25in,clip,keepaspectratio]{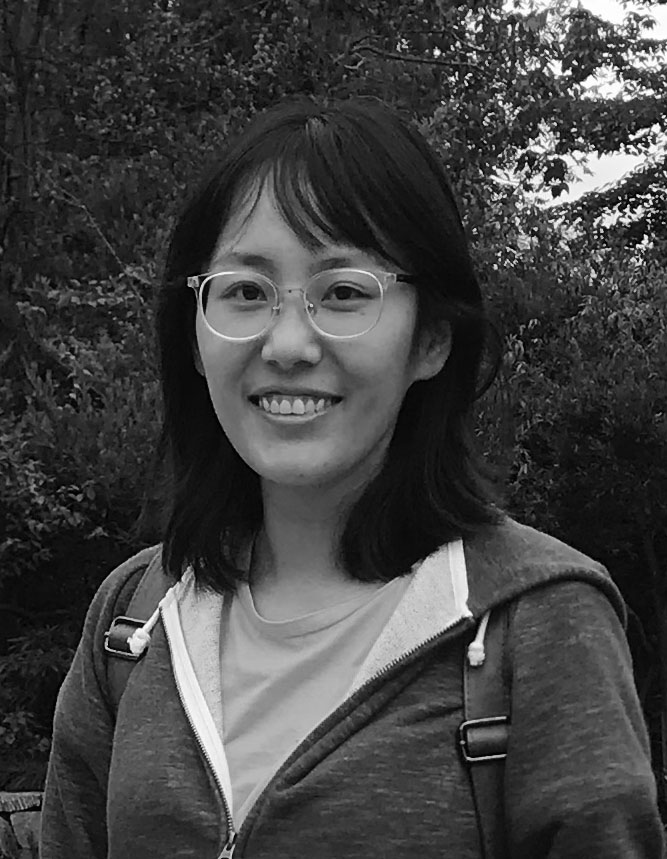}}]{Mengqi Zhu} received her M.Sc. degree from the Beijing Institute of Technology (BIT), Beijing, China, in 2017. She is currently an engineer in the Information and Control Technology Department at the North China Vehicle Research Institute, while also pursuing a Ph.D. at BIT. Her research interests focus on image processing and intelligent perception.
\end{IEEEbiography}

\begin{IEEEbiography}[{\includegraphics[width=1in,height=1.25in,clip,keepaspectratio]{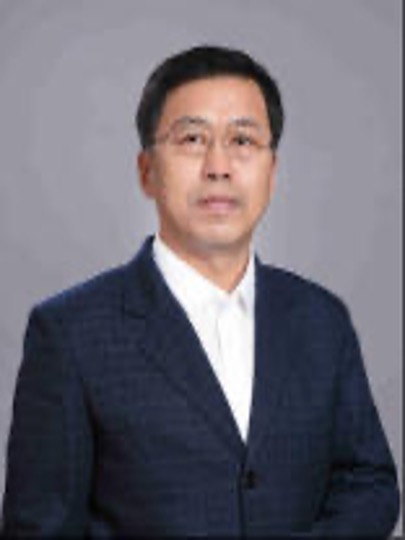}}]{LIQIANG SONG} received the Ph.D. degree from the University of Chinese Academy of Sciences, Beijing, China, in July 2009. From July 2009 to May 2024, he was with the National Astronomical Observatories of the Chinese Academy of Sciences (NAOC), where he successively served as subsystem manager for the Five-hundred-meter Aperture Spherical radio Telescope (FAST), assistant commissioning group leader, deputy director of the Engineering Office, and leader of the Structures Group. His research interests include radio astronomy instrumentation, telescope commissioning, and structural design of large-scale radio telescopes.
\end{IEEEbiography}

\begin{IEEEbiography}[{\includegraphics[width=1in,height=1.25in,clip,keepaspectratio]{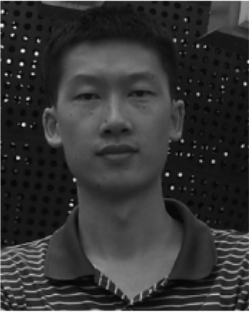}}]{Yuqiang Fang} received the Ph.D. degree in control science and engineering from the National University of Defense Technology, Changsha, China, in 2015. He is currently an Associate Professor with Space Engineering University, Beijing, China. His research interests include machine learning, computer vision, and data mining.
\end{IEEEbiography}

\begin{IEEEbiography}[{\includegraphics[width=1in,height=1.25in,clip,keepaspectratio]{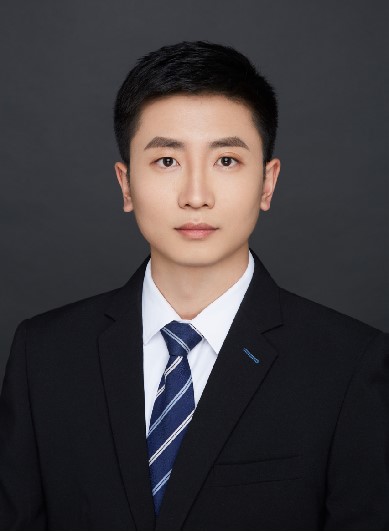}}]{Jianan Li} is currently an assistant professor at School of Optics and Photonics, Beijing Institute of Technology, Beijing, China, where he received his B.S. and Ph.D. degree in 2013 and 2019, respectively. From July 2015 to July 2017, he worked as a joint training Ph.D. student at National University of Singapore. From October 2017 to April 2018, he worked as an intern at Adobe Research. His research interests mainly include computer vision and real-time image/video processing.
\end{IEEEbiography}

\vfill

\end{document}